\documentclass[10pt,twocolumn,letterpaper]{article}

\usepackage{cvpr}              

\usepackage[table]{xcolor}

\definecolor{customlinkcolor}{HTML}{2C2F8F}
\definecolor{cvprblue}{rgb}{0.21,0.49,0.74}
\usepackage[pagebackref,breaklinks,colorlinks,allcolors=cvprblue]{hyperref}
\usepackage{booktabs}
\usepackage{array}
\usepackage{multirow}
\usepackage{algorithm}
\usepackage{algpseudocode}
\usepackage[accsupp]{axessibility}
\def\paperID{*****} 
\def\confName{CVPR}
\def\confYear{2026}

\title{Seen-to-Scene: Keep the Seen, Generate the Unseen for Video Outpainting}

\author{Inseok Jeon$^{1}$\quad Minhyeok Lee$^{1}$\quad Seunghoon Lee$^{1}$\quad Minseok Kang$^{1}$\quad Suhwan Cho$^{2}$\quad Sangyoun Lee$^{1}$\quad \vspace{0.3cm}\\
$^1$~Yonsei University\quad $^2$~GenGenAI\vspace{0.2cm}\\
{\fontsize{10.0}{10.0}\selectfont
\textit{\href{https://inseokjeon.github.io/seen_to_scene}{\textcolor{customlinkcolor}{\nolinkurl{https://inseokjeon.github.io/seen_to_scene}}}}}}

\begin{document}
\twocolumn[{
	\renewcommand\twocolumn[1][]{#1}
	\maketitle
	\begin{center}
		\centering
		\captionsetup{type=figure}
		\vspace{-7mm}
		\includegraphics[width=1\linewidth]{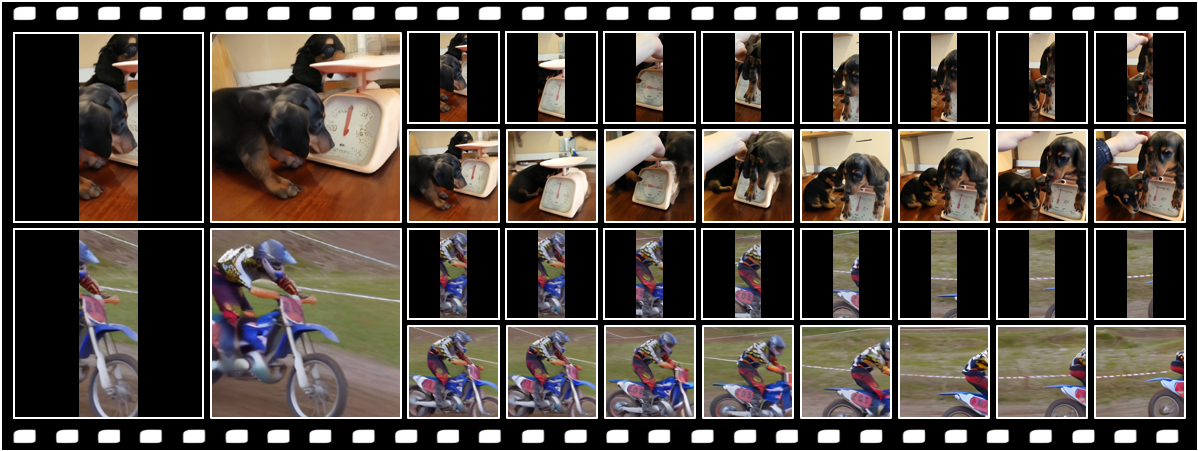}
		\vspace{-7mm}
		\caption{\textbf{Seen-to-Scene} is a novel framework that unifies propagation-based and generation-based paradigms for video outpainting. It maintains strong spatio-temporal consistency across frames even under challenging conditions such as camera motion, dynamic objects and complex motion patterns. The official implementation is released at \href{https://your-code-url}{\textcolor{customlinkcolor}{\textit{\nolinkurl{https://github.com/InSeokJeon/Seen_to_Scene}}}}.}
		\label{figure1}
	\end{center}
}]

\begin{abstract}
Video outpainting aims to expand the visible content of a video beyond the original frame boundaries while preserving spatial fidelity and temporal coherence across frames. Existing methods primarily rely on large-scale generative models, such as diffusion models. However, generation-based approaches suffer from implicit temporal modeling and limited spatial context. These limitations lead to intra-frame and inter-frame inconsistencies, which become particularly pronounced in dynamic scenes and large outpainting scenarios. To overcome these challenges, we propose \textbf{Seen-to-Scene}, a novel framework that unifies propagation-based and generation-based paradigms for video outpainting. Specifically, Seen-to-Scene leverages flow-based propagation with a flow completion network pre-trained for video inpainting, which is fine-tuned in an end-to-end manner to bridge the domain gap and reconstruct coherent motion fields. To further improve the efficiency and reliability of propagation, we introduce a reference-guided latent propagation that effectively propagates source content across frames. Extensive experiments demonstrate that our method achieves superior temporal coherence and visual realism with efficient inference, surpassing even prior state-of-the-art methods that require input-specific adaptation.
\end{abstract}

\section{Introduction}
\label{sec:intro}
As short-form and multi-platform video content grows rapidly, video production and consumption are expanding across a wide range of devices and displays. In practice, most videos are typically captured at fixed aspect ratios such as 16:9 and 9:16, while modern displays offer diverse aspect ratios and sizes. This mismatch often leads to black bars or undesired cropping, which remains a practical challenge in modern video editing and content creation.

Video outpainting, which generates visual content beyond the original frame boundaries across the video sequence, has emerged as a promising solution. Recently, a wide variety of methods have been proposed, and they can be categorized into two paradigms: \textbf{(i)} propagation-based approaches and  \textbf{(ii)} generation-based approaches.

Propagation-based approaches complete the outpainting regions by warping neighboring frames using optical flow and propagating observable content into missing areas.
Such propagation methods have been widely used in video inpainting~\cite{li2022towards, zhang2022inertia, zhang2022flow, zhou2023propainter, cho2025elevating, lee2025video}, where masked regions are restored by propagating reliable spatial and temporal information from neighboring frames. Building upon propagation-based approaches originally designed for video inpainting, Dehan~\cite{dehan2022complete} introduced a propagation-based framework for video outpainting. However, applying these propagation strategies to video outpainting leads to several limitations. First, dense pixel-level warping and sequential flow composition across intermediate frames demand substantial computation, as the cost scales with both spatial resolution and the number of frames. Unlike video inpainting, which fills in missing regions within a fixed spatial resolution, video outpainting inherently requires spatial expansion beyond the original frame boundaries. As a result, propagation costs increase substantially in video outpainting. Second, existing methods rely on flow-completion networks trained for video inpainting, which introduces a domain gap when applied to video outpainting. These networks are optimized to estimate motion within localized masked regions and therefore struggle to generalize to the large spatial expansions required in outpainting. As the distance from the source region increases, the predicted flows become progressively unstable, resulting in flow misalignment and visible artifacts. Finally, propagation alone cannot synthesize non-observable regions and therefore necessitates generative models or auxiliary networks to fill the missing areas.

\vspace{-0.27mm}
Motivated by these limitations and the success of generative models, recent studies~\cite{fan2023hierarchical, wang2024your, chen2024follow, zhong2025outdreamer, lidynamic, murakawa2026m3ddm+, pan2026globalpaint, yu2025unboxed} have shifted toward generation-based approaches. Leveraging diffusion models~\cite{ho2020denoising} with powerful  generative capability, they have achieved significant progress. However, even with these advances, generation-based approaches still face key challenges. First, generative models cannot ensure consistent preservation of source content over time, as diffusion-based models rely on implicit temporal correlations. Such modeling maintains source content between adjacent frames, but its temporal correlations break down over longer intervals, causing inter-frame inconsistencies and temporal discontinuities. Second, limited spatial cues restrict global scene understanding, forcing reliance on learned generative priors. These models often hallucinate content, resulting in intra-frame inconsistencies and structural artifacts. Overall, achieving spatio-temporal coherence remains an open challenge for video outpainting.

\vspace{-0.27mm}
To overcome the limitations of these two paradigms, we propose \textbf{Seen-to-Scene}, a novel propagation-based video diffusion framework for video outpainting. At its core, Seen-to-Scene unifies propagation and generation by combining the source content preservation of propagation with the generative capability of diffusion models. The integration of latent propagation into pre-trained video diffusion models~\cite{blattmann2023stable} contributes in three key aspects. \textbf{(i) Implicit Latent Diversity}: Latent propagation introduces variations in spatial structure and temporal dynamics across frames. Such variations inherently expose the model to diverse spatial layouts and motion patterns, improving its generalization across a wide range of visual scenes and spatial expansion ratios. \textbf{(ii) Enhancing Spatial Reasoning}: Propagated latents provide additional spatial cues beyond frame boundaries. Such guidance promotes global structural awareness and coherent spatial layout reasoning, mitigating hallucinated structures and improving spatial coherence. \textbf{(iii) Improving Temporal Consistency}: Explicit propagation across frames encourages the model to preserve observable regions while generating non-observable content. This mechanism further strengthens temporal coherence and structural consistency throughout the sequence.

\begin{figure}[t]
    \centering
    \includegraphics[width=1\columnwidth]{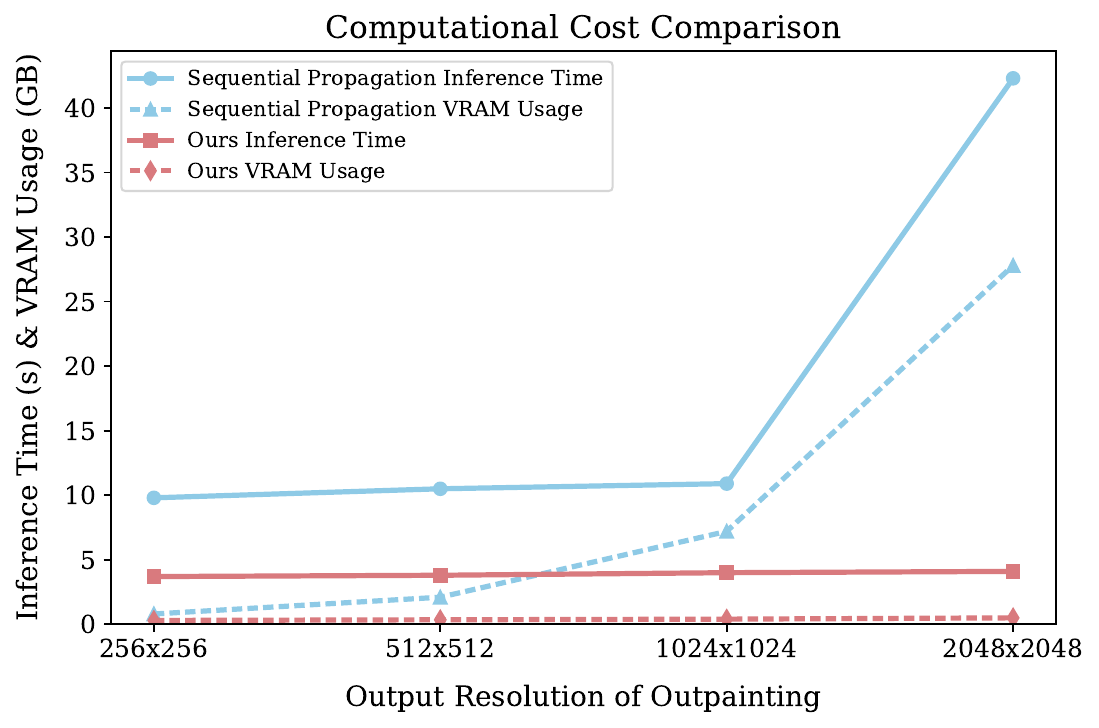}
    \vspace{-0.8cm}
    \caption{Computational cost comparison with a SOTA propagation-based model~\cite{cho2025elevating}. Dense and sequential propagation incurs significantly higher inference time and VRAM usage.}
    \vspace{-0.7cm}
    \label{fig:computational_cost}
\end{figure}

To effectively integrate propagation into a generative model, several practical challenges must be addressed. Conventional propagation schemes incur significant computational overhead due to dense warping in pixel space and sequential flow composition across frames. We alleviate this issue by introducing a reference-guided latent propagation that identifies content-rich reference frames via inter-frame correlation within a sliding temporal window and performs propagation directly in the latent space. Since propagation is directly integrated into the generation process, misaligned propagated latents can adversely affect subsequent generation. We also design a lightweight refinement module to mitigate propagation artifacts and reduce flow dependency. Furthermore, we observe that existing flow completion networks trained for video inpainting lead to a non-trivial domain gap when applied to video outpainting. We therefore adapt the flow completion network to video outpainting, yielding robust motion estimates for propagation.

\begin{figure*}[t]
\centering
\includegraphics[width=1\linewidth]{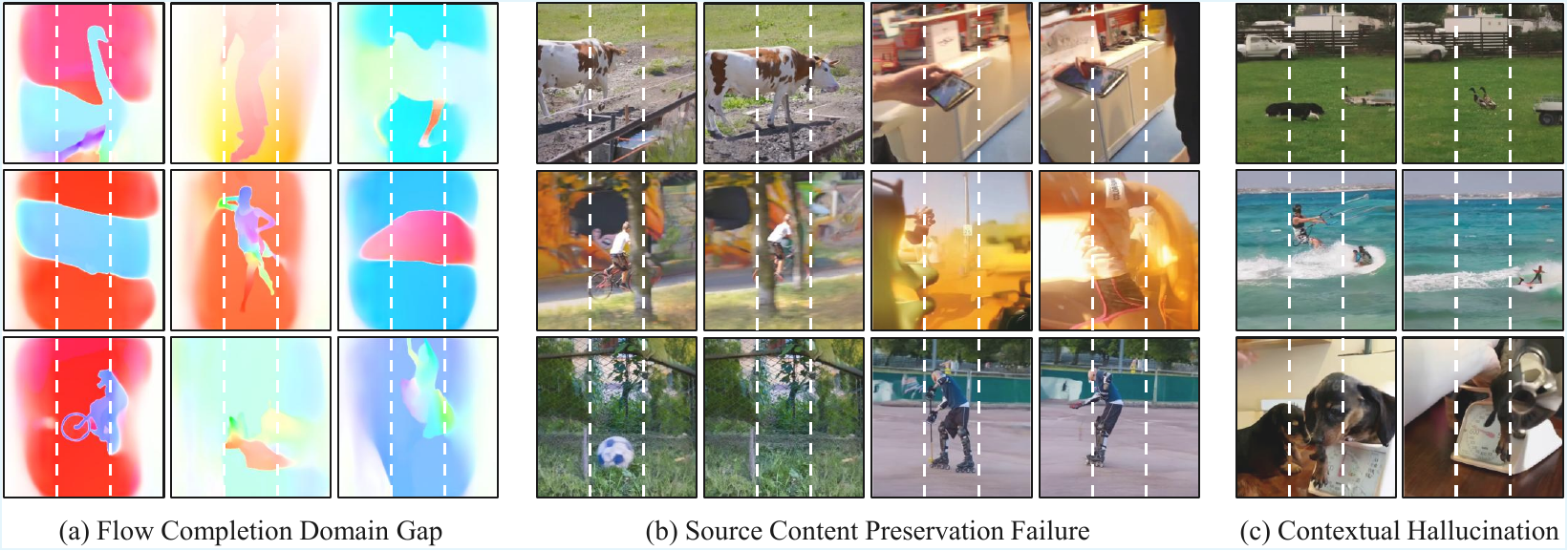}
\vspace{-7mm}
\caption{\textbf{Key limitations of existing video outpainting approaches.}
(a) \textit{Flow completion domain gap}: Flow completion networks trained for video inpainting struggle to generalize to video outpainting due to the substantially larger missing regions involved. Consequently, the predicted flow often suffers from bleeding around dynamic objects and fails to provide reliable motion estimates for regions far from the original content.
(b) \textit{Source content preservation failure}: Generation-based approaches often fail to preserve the original source content across frames when spatially extending video frames. As a result, visual content may disappear or fail to appear consistently in the generated regions, leading to temporal inconsistency throughout the video sequence.
(c) \textit{Contextual hallucination}: Due to the limited contextual cues available for video outpainting, generation-based methods may hallucinate semantically irrelevant or unrealistic content in the outpainting regions. Such hallucinations introduce visual artifacts and intra-frame inconsistency.}
\vspace{-5mm}
\label{figure2}
\end{figure*}

\vspace{0.2mm}
Our main contributions can be summarized as follows:
\vspace{-0.1mm}
\begin{itemize}
    \item  We propose Seen-to-Scene, a novel propagation-based video diffusion model that integrates propagation and generation for video outpainting to preserve observed content while naturally generating unseen regions.
    \item We introduce an efficient reference-guided latent propagation tailored for video outpainting that retrieves informative frames based on cross-frame similarity, enabling propagation without sequential frame-wise composition.
    \item To improve the alignment of propagated latents, we design a refinement module and conduct the first analysis of the domain gap in flow completion for video outpainting.
    \item Despite not requiring input-specific adaptation, Seen-to-Scene achieves superior visual fidelity and temporal coherence with efficient inference on standard outpainting benchmarks composed of diverse real-world videos.
\end{itemize}

\section{Related Work}
\label{sec:formatting}

\textbf{Flow-based propagation.} Videos inherently exhibit strong spatio-temporal continuity, as adjacent frames share highly correlated structures and motion patterns. This motivates the use of optical flow to explicitly capture pixel-wise motion across frames. To preserve source content, flow-based propagation typically consists of three stages: flow estimation, flow completion, and flow-guided warping. Prior methods have improved these components individually or jointly. DFVI~\cite{xu2019deep} completes optical flow fields using a Deep Flow Completion Network (DFC-Net) and propagates known pixels along the flow to fill missing areas. E2FGVI~\cite{li2022towards} jointly learns flow completion and propagation. ProPainter~\cite{zhou2023propainter} enhances propagation through dual-domain integration. RGVI~\cite{cho2025elevating} introduces a one-shot pixel pulling strategy to mitigate error accumulation. Although these approaches have achieved notable success in video inpainting, they are inherently limited when extended to video outpainting due to the lack of observable context and the large spatial extent of missing regions. In addition, their pixel-level warping relies on sequential flow composition across intermediate frames, resulting in higher computational cost as spatial resolution and video length increase. These limitations are illustrated in Fig.~\ref{fig:computational_cost} and Fig.~\ref{figure2} \textit{(a)}.

\noindent \textbf{Video outpainting.} Unlike video inpainting, which fills in missing regions within a fixed resolution, video outpainting expands the spatial canvas by generating new content beyond the original frame boundaries. Previous works on video outpainting can be categorized into zero-shot and one-shot methods based on inference strategies. 

Zero-shot methods generate spatially extended content directly from the input video without any fine-tuning on the specific video. Dehan~\cite{dehan2022complete} estimates optical flow to warp background content from adjacent frames and fills the remaining regions using an image completion network. M3DDM~\cite{fan2023hierarchical} trains a 3D diffusion model on large-scale video datasets with mask modeling. Follow-Your-Canvas~\cite{chen2024follow} distributes the generation process across spatial windows to overcome GPU memory constraints. However, these approaches often suffer from limited visual fidelity, temporal inconsistency, and blurry results. In particular, Li et al.~\cite{lidynamic} propose a flow completion network for video outpainting, yet their method directly conditions the generation process on completed flow, leading to flow dependency.

In contrast, one-shot methods perform fine-tuning on each specific input video before generating new content. MOTIA~\cite{wang2024your} performs input-specific adaptation to capture sequence-specific appearance and motion patterns. Unboxed~\cite{yu2025unboxed} expands static scenes via 3D Gaussian Splatting and utilizes diffusion models for dynamic-region synthesis and temporal coherence. Although these methods achieve strong performance, they incur substantial inference overhead from sequence-specific adaptation and often produce repetitive or irrelevant content in the outpainting regions. Fig.~\ref{figure2} illustrates several limitations of existing methods.

Unlike prior approaches that handle propagation and generation separately, we propose a unified framework that jointly exploits their complementary strengths within a single end-to-end framework. By integrating propagation-driven source preservation with diffusion-based content generation, our method preserves original structures while providing additional spatial cues during generation. Despite operating in a zero-shot manner, our framework outperforms existing one-shot approaches while achieving faster inference with reduced computational resources.

\begin{figure*}[t]
\centering
\includegraphics[width=1\linewidth]{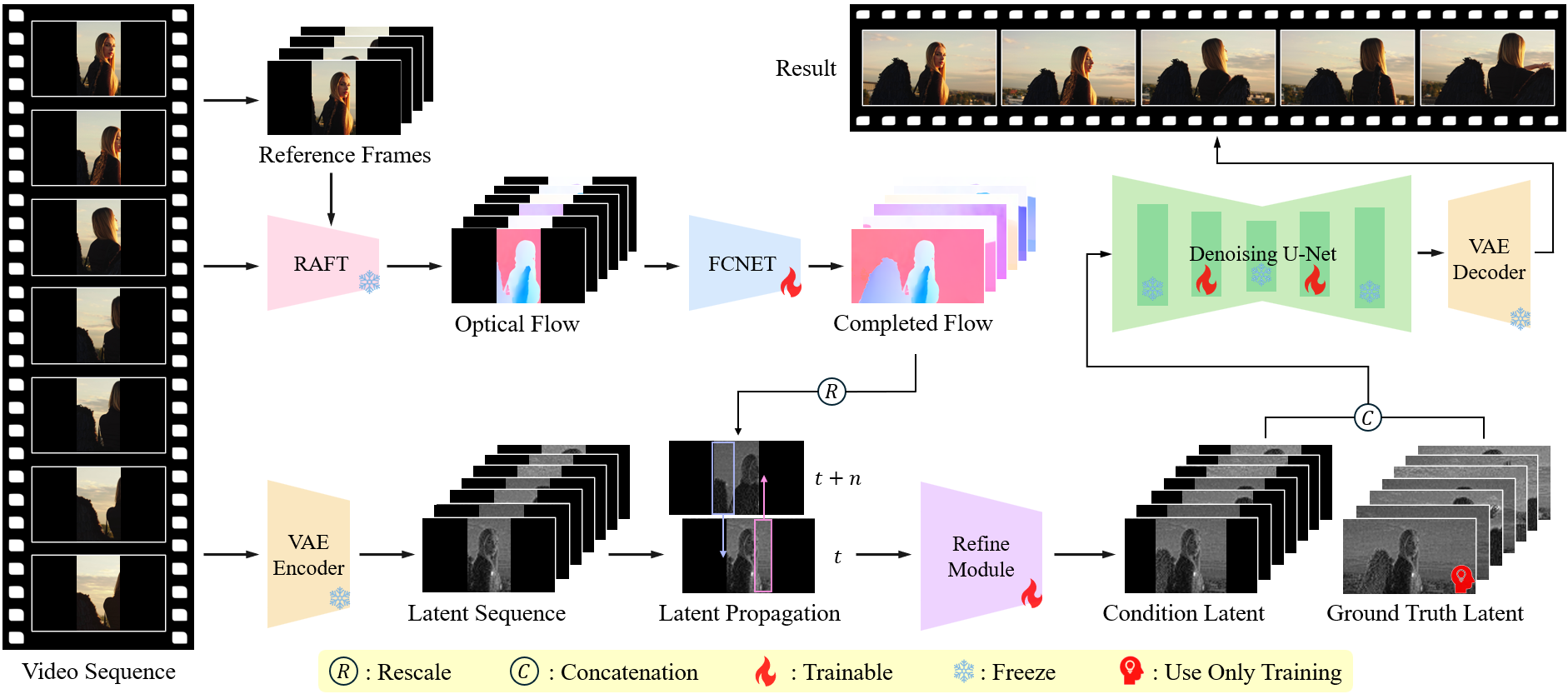}
\vspace{-3mm}
\caption{\textbf{Overview of the proposed Seen-to-Scene framework.}
Seen-to-Scene unifies propagation and generation for coherent video outpainting. Given an input video, reference frames are selected and optical flow is estimated using RAFT~\cite{teed2020raft}, followed by flow completion with FCNet~\cite{zhou2023propainter}. The input frames are encoded into latent representations via a pre-trained VAE encoder. Using the completed flow, we perform reference-guided latent propagation to preserve source content across frames. A lightweight refinement module is then applied to mitigate propagation artifacts. The refined latents serve as conditions for a diffusion-based generation process, where a denoising U-Net synthesizes the outpainted content in latent space. The final video frames are reconstructed using a VAE decoder.}
\vspace{-3mm}
\label{figure3}
\end{figure*}

\section{Preliminary}
\noindent\textbf{Diffusion Model.}
A typical diffusion model consists of three components: a pre-trained VAE encoder~\cite{kingma2013auto} that maps each video frame into a latent code, a denoising network (e.g., a 3D-UNet or Transformer~\cite{rombach2022high}) that models the diffusion process in the latent space, and a VAE decoder that reconstructs video frames from the latent representations.

Given a video consisting of $N$ frames, the encoder maps each frame into a latent code $\mathbf{z}_i$, forming the latent sequence $\mathbf{Z}_0 = \{\mathbf{z}_i\}_{i=1}^{N}$. In the forward diffusion process, Gaussian noise is progressively added to $\mathbf{Z}_0$ over $T$ time steps. With a pre-defined noise schedule $\{\beta_t\}_{t=1}^{T}$, we define $\alpha_t = 1 - \beta_t$ and $\bar{\alpha}_t = \prod_{i=1}^{t}\alpha_i$, where $\alpha_t$ controls the noise level at step $t$, and $\bar{\alpha}_t$ represents the cumulative noise scaling factor that determines the relative contribution of the original signal in $\mathbf{Z}_t$. The noisy latent $\mathbf{Z}_t$ at time step $t$ can be written as \begin{equation}
\mathbf{Z}_t = \sqrt{\bar{\alpha}_t}\,\mathbf{Z}_0 
+ \sqrt{1 - \bar{\alpha}_t}\,\boldsymbol{\epsilon}, 
\qquad 
\boldsymbol{\epsilon} \sim \mathcal{N}(\mathbf{0}, \mathbf{I}).
\end{equation} where $\boldsymbol{\epsilon}$ denotes Gaussian noise sampled from a standard normal distribution. The denoising network $\boldsymbol{\epsilon}_\theta$ is trained to predict the injected noise given $(\mathbf{Z}_t, t)$ by minimizing the standard diffusion training objective:
\begin{equation}
\label{objective_function}
\mathcal{L}
= \mathbb{E}_{t,\,\boldsymbol{\epsilon}\sim\mathcal{N}(\mathbf{0},\mathbf{I})}
\big\|\,\boldsymbol{\epsilon}
- \boldsymbol{\epsilon}_{\theta}(\mathbf{Z}_t, t)\,\big\|^{2}.
\end{equation}
In the inference stage, the denoising network iteratively removes noise from $\mathbf{Z}_T$ to obtain the clean latent sequence $\mathbf{Z}_0$. The resulting latent sequence is then decoded by the VAE decoder to reconstruct the final video frames.

\section{Method}
We present \textbf{Seen-to-Scene}, a novel video outpainting framework that integrates propagation and generation within a unified architecture. We first formulate the problem in Sec.~\ref{sec:problem_formulation}. We then describe the reference-guided latent propagation mechanism in Sec.~\ref{latent_propagate}. Finally, we describe the diffusion-based generation framework in Sec.~\ref{video diffusion}.
\subsection{Problem Formulation}
\label{sec:problem_formulation}
Given a video sequence $\{I_i\}_{i=1}^{N}$ consisting of $N$ frames, where each frame $I_i \in \mathbb{R}^{3 \times h \times w}$, our goal is to expand each frame beyond its original spatial canvas. Specifically, we aim to generate an expanded sequence $\{\hat{I}_i\}_{i=1}^{N}$, where $\hat{I}_i \in \mathbb{R}^{3 \times H \times W}$ with $H > h$ and $W > w$, while preserving both intra-frame and inter-frame consistency. The overall pipeline of the proposed framework is illustrated in Fig.~\ref{figure3}.

\subsection{Latent Propagation}
\label{latent_propagate}

Preserving source content across frames is essential for maintaining both intra-frame and inter-frame consistency. Observable content can be propagated across frames to provide reliable spatial cues that partially extend the outpainting regions. This insight motivates our latent propagation, which propagates visual information in latent space to guide the generation of coherent outpainting regions. We now describe our reference-guided latent propagation in detail.

\vspace{1.5mm}
\noindent \textbf{Reference Frame Selection.} 
Video sequences inherently exhibit strong temporal continuity, where adjacent frames share highly overlapping visual content. To avoid redundant information while providing informative context for the outpainting regions, we select reference frames based on structural correlation. For an input sequence $\{I_i\}^{N}_{i=1}$, we define a temporal window of size $m$ and construct a global reference chain $\mathcal{R} = [r_1, r_2, \dots, r_L]$, initialized with the first frame $r_1 = I_1$. At step $l$, given the current reference index $r_l$, we define the candidate index set within the temporal window as $\mathcal{C}(r_{l+1}) =\{r_l+1,\, r_l+2,\, \ldots,\, \min(r_l + m,\; N) \}$. For each candidate frame $I_j$ with $j \in \mathcal{C}(r_{l+1})$, we measure the structural correlation between $I_{r_l}$ and $I_j$ using the structure component of the Structural Similarity Index (SSIM)~\cite{wang2004image}. To suppress illumination and color variations and focus purely on structural similarity, we compute the structure term of SSIM on gray-scale images as follows:
\begin{equation}
s(r_l, j) \;=\; 
\mathrm{SSIM}_{\text{structure}}\!\big(
\mathrm{gray}(I_{r_l}),\,
\mathrm{gray}(I_j)
\big),
\end{equation}
where $s(r_l, j)$ denotes the structural correlation score between frames $I_{r_l}$ and $I_j$.
We select the next reference frame by minimizing the score within the candidate set:
\begin{equation}
r_{l+1} \;=\; 
{\arg\min}_{\,j \in \mathcal{C}(r_{l+1})} \; s(r_l, j).
\end{equation}
Intuitively, selecting the frame with the lowest spatial similarity encourages the reference chain to include frames that contain rich contextual information while avoiding redundant content from temporally adjacent frames. After selecting $r_{l+1}$, the temporal window is advanced to the new reference index and the same procedure is iteratively applied to extend the reference chain. When the number of candidates becomes smaller than the window size $m$, the final frame $I_N$ is appended to $\mathcal{R}$ to complete the reference sequence.

\vspace{1.5mm}
\noindent \textbf{Flow Estimation and Completion.}
To preserve source content via flow-based propagation, we estimate optical flow using RAFT~\cite{teed2020raft} between each frame $I_i$ and its nearest past $r^{-}(i)$ and future $r^{+}(i)$ reference frames, defined as:
\begin{equation}
\begin{aligned}
F_{i\to r^{-}(i)} &= \mathrm{RAFT}(I_i, I_{r^{-}(i)}) 
\quad {\scriptstyle r^{-}(i)=\max\{r_l : r_l \le i\}}, \\
F_{i\to r^{+}(i)} &= \mathrm{RAFT}(I_i, I_{r^{+}(i)}) 
\quad {\scriptstyle r^{+}(i)=\min\{r_l : r_l \ge i\}}.
\end{aligned}
\end{equation}
On the expanded canvas, we define a binary outpainting mask $\{M_i\}_{i=1}^{N}$ with $M_i \in \mathbb{R}^{1 \times H \times W}$, where $M_i(x)=1$ denotes the outpainting region outside the original frame boundary. Each estimated flow is mapped onto the expanded canvas. The regions indicated by the outpainting mask are treated as missing areas to be completed. To complete the missing flows in the outpainting regions, we adopt a pre-trained flow completion network~\cite{zhou2023propainter}. However, existing flow completion models for video inpainting are typically specialized for small and spatially contiguous missing regions. When applied to large outpainting masks, these models often exhibit a domain gap, resulting in spatially diffuse or inconsistent flow estimates. To mitigate this domain gap, we jointly fine-tune the pre-trained flow completion network within our end-to-end pipeline using outpainting masks as supervision for the missing regions. This joint optimization allows the network to adapt to the large and open-ended missing regions in outpainting, thereby improving flow reliability beyond the original frame boundaries. As a result, we obtain completed optical flows $\{\hat{F_i}\}_{i=1}^{N-1}$ between each frame $I_i$ and its reference frames $I_{r(i)}$.

\noindent \textbf{Reference Guided Latent Propagation.}
Each frame $I_i$ is encoded by the VAE encoder into a latent code $Z_i \in \mathbb{R}^{c' \times h/s \times w/s}$, where $s$ denotes the spatial downsampling factor. Since the VAE encoder preserves spatial correspondence, the completed flows $\{\hat{F_i}\}_{i=1}^{N-1}$ in pixel space and the outpainting masks $\{M_i\}_{i=1}^{N}$ are downscaled by the same factor $s$. Each latent $Z_i$ is placed on the expanded latent canvas according to the mask. To fill in the outpainting regions, we propagate latent codes from reference frames using the completed flows and one-shot pulling strategy~\cite{cho2025elevating}. For each target latent $Z_i$, latent codes are aggregated only along the pre-defined reference chain instead of across all frames. Specifically, starting from the nearest past and future reference $r^{-}(i)$ and $r^{+}(i)$, the corresponding reference latents are first propagated to the target latent ${Z}_i$ via grid-based warping using the completed flows $F_{i\to r^{-}(i)}$ and $F_{i\to r^{+}(i)}$. We further extend the propagation range by recursively composing reference-to-reference flows, allowing temporally distant references to contribute without requiring direct flow estimation between all frame pairs. For references ${r_t}$ ordered in time, the accumulated flows toward the target latent $Z_i$ are defined as follows: 
\begin{equation}
\begin{aligned}
F_{i \to r^{+}_{t+1}(i)}
&= F_{i\to r^{+}_t(i)} 
  + \mathcal{W}\!\big(F_{r^{+}_t(i)\to r_{t+1}^{+}(i)}\,;\ F_{i\to r^{+}_t(i)}\big), \\[4pt]
F_{i \to r^{-}_{t-1}(i)}
&= F_{i\to r^{-}_t(i)} 
  + \mathcal{W}\!\big(F_{r^{-}_{t-1}(i)\to r_{t}^{-}(i)}\,;\ F_{i\to r^{-}_t(i)}\big).
\end{aligned}
\end{equation}
Using the accumulated reference flows, we warp reference latents to the target frame under the corresponding source masks $\dot{M_i}=1-M_i$. This enables direct long-range propagation along the reference chain instead of sequential propagation, reducing the accumulation of warping errors.

\noindent \textbf{Latent Alignment.}
Through latent propagation, each latent aggregates globally observable visual content across the video sequence. To further reduce the dependency on completed flow and mitigate local misalignments, we introduce a lightweight refinement module that refines latent alignment using appearance cues rather than motion information. The refinement module takes the propagated latent, the original latent, the outpainting mask, and the propagated mask as inputs. It refines locally misaligned structures by predicting residual sampling offsets and adaptive modulation weights that selectively adjust latent alignment within uncertain regions. Alignment is performed bidirectionally, and the two direction-specific latents are fused using a shallow convolution layer to produce the final refined latent representation. Additional details, including the architecture of the refinement module and pseudo code for the latent propagation procedure, are provided in Appendix C.

\subsection{Video Diffusion Model}
\label{video diffusion}
After latent propagation, the outpainting regions in the latents are partially filled with content propagated from source regions across frames. This propagated content serves as conditioning for the subsequent diffusion process.

\vspace{0.8mm}
\noindent \textbf{Training Stage.} For diffusion training, we encode the ground-truth video into latent codes $\{\dot{Z}_i\}_{i=1}^{N}$ using a pre-trained VAE encoder. The propagated latents $\{\tilde{Z}_i\}_{i=1}^{N}$, which contain source content partially propagated through reference-guided latent propagation, are concatenated with $\{\dot{Z}_i\}_{i=1}^{N}$ along the channel dimension to form the input to the denoising network. Gaussian noise is then added to the concatenated latents according to the diffusion timestep to obtain the noisy latents for training. The noisy latents are fused using a $1 \times 1$ convolution layer and fed into the 3D U-Net backbone. To leverage the strong spatio-temporal priors learned from large-scale video data, we fine-tune a pre-trained video diffusion model~\cite{blattmann2023stable} for the video outpainting task. Specifically, all convolutional and spatial attention layers are frozen to preserve the strong spatial priors learned during large-scale pre-training, while only the temporal transformer blocks remain trainable. The temporal blocks leverage motion-aware propagation-conditioned latents to aggregate temporal context and perform temporal reasoning across frames. The denoising network is then optimized for outpainting using the objective defined in Eq.~\ref{objective_function}.

\vspace{0.8mm}
\noindent \textbf{Inference Stage.} During inference, the ground-truth latent used during training is replaced with Gaussian noise sampled from a standard normal distribution. This noise serves as the initial noise sample for the generation process. The propagated latents obtained from the reference-guided latent propagation are concatenated with the noise latent along the channel dimension. This conditioning provides contextual guidance that enables the diffusion model to generate coherent and plausible outpainting content. Unlike existing state-of-the-art approaches~\cite{wang2024your, chen2024follow}, our approach does not require text prompts or additional conditioning signals during inference. The concatenated latents are then fed into the 3D U-Net to predict the outpainted latents. Finally, the VAE decoder reconstructs the final video frames. As a result, our method performs inference in a zero-shot manner.

\vspace{-3mm}
\section{Experiments}

\subsection{Experiment Setup}
\label{training setup}

\noindent \textbf{Implementation Details.} For latent propagation, we set the window size to $m=4$. For diffusion training, we construct 25-frame video clips. The model is trained using Adam~\cite{kingma2014adam} with an initial learning rate of $1\times10^{-5}$ for 100K iterations, with a batch size of 1 on two NVIDIA RTX A6000 GPUs.

\noindent \textbf{Training Datasets.} Previous diffusion-based video outpainting methods are trained on large-scale or private datasets containing millions of video samples~\cite{chen2024panda}. In contrast, our model is trained using only 100K video samples from the publicly available YouTube-VOS training set~\cite{xu2018youtube}.

\noindent \textbf{Evaluation Datasets.} To align with existing benchmarks, we evaluate our method on the DAVIS~\cite{perazzi2016benchmark} and YouTube-VOS~\cite{xu2018youtube} datasets. 
For DAVIS 2017, we test on 90 video sequences. The evaluation protocol for YouTube-VOS is not explicitly standardized in prior works~\cite{chen2024follow, yu2025unboxed}, which often rely on selected high-quality subsets.  To avoid potential selection bias, we randomly sample 60 videos from the official test set. Following the standard protocol, we report horizontal outpainting results under mask ratios of 0.25 and 0.66 for all test videos. The complete list of sampled videos is provided in Appendix E to ensure reproducibility.

\vspace{1mm}
\noindent \textbf{Evaluation Metrics.} We evaluate our method using standard quantitative metrics, including PSNR, SSIM~\cite{wang2004image}, LPIPS~\cite{zhang2018unreasonable}, and FVD~\cite{unterthiner2018towards}. Fréchet Video Distance (FVD) serves as the principal metric for video outpainting, as it captures both perceptual realism and temporal consistency by measuring the distributional distance between the generated and original videos. All evaluation metrics are computed following the same protocol as prior works~\cite{chen2024follow, wang2024your}.

\subsection{Quantitative Evaluation} Table~\ref{table1} reports the quantitative comparison with existing video outpainting methods on the DAVIS and YouTube-VOS datasets. Our method outperforms prior approaches across all evaluation metrics on both datasets. In particular, on DAVIS, our method achieves the best performance with a PSNR of 21.95, LPIPS of 0.141, SSIM of 0.783, and FVD of 218.8, surpassing previous methods by a clear margin. Similarly, on YouTube-VOS, our method outperforms all baselines across all four metrics, achieving improvements of 6.7\% in FVD and 19.7\% in LPIPS, indicating superior perceptual quality and temporal stability. These results demonstrate the effectiveness of our reference-guided latent propagation in preserving source content while maintaining coherent spatio-temporal structures during video outpainting. Notably, all results are achieved without any scene-specific adaptation, highlighting the strong generalization capability of Seen-to-Scene across diverse video scenarios.

\subsection{Qualitative Evaluation} Fig.\ref{figure4} illustrates the qualitative comparison between our method and existing approaches. Although M3DDM~\cite{fan2023hierarchical} introduces masked modeling for data augmentation, it often produces blurry content or fails to generalize when generating regions distant from the observed area. MOTIA~\cite{wang2024your}, which performs test-time adaptation to capture input-specific patterns, frequently leads to repetitive objects or background artifacts. Follow-Your-Canvas~\cite{chen2024follow} occasionally generates hallucinated structures unrelated to the original video content. Fundamentally, these methods fail to preserve the source content observed in the original sequence. In contrast, our method faithfully preserves the source content while maintaining strong spatio-temporal consistency, producing coherent outpainting results even in challenging scenarios. Additional results are provided in Appendix E.

\newcolumntype{P}[1]{>{\centering\arraybackslash}p{#1}}
\begin{table*}[t!]
\centering 
\small
\caption{\textbf{Quantitative evaluation on DAVIS and YouTube-VOS.} Following prior benchmarks, the source videos with different aspect ratios are outpainted to 256×256 resolution. Gray rows indicate methods requiring input-specific adaptation.}
\vspace{-3mm}
\begin{tabular}{p{3.2cm}P{2.1cm}P{1.0cm}P{1.0cm}P{1.0cm}P{1.0cm}P{1.0cm}P{1.0cm}P{1.0cm}P{1.0cm}}

\toprule
\multirow{2}{*}{\centering Method} &
\multirow{2}{*}{\centering Publication} &
\multicolumn{4}{c}{DAVIS} &
\multicolumn{4}{c}{YouTube-VOS} \\
\cmidrule(lr){3-6}\cmidrule(lr){7-10}
& & PSNR$\uparrow$ & LPIPS$\downarrow$ & SSIM$\uparrow$ & FVD$\downarrow$
  & PSNR$\uparrow$ & LPIPS$\downarrow$ & SSIM$\uparrow$ & FVD$\downarrow$ \\
\midrule

Dehan~\cite{dehan2022complete}                          & CVPRW'22  & 17.96 & 0.233 & 0.627 & 363.1 & - & - & - & - \\
M3DDM~\cite{fan2023hierarchical}                   & ACMMM'23  & 20.26 & 0.204 & 0.708 & 300.0 & 20.93 & 0.186 & \underline{0.704} & 266.2 \\
\rowcolor{gray!20} MOTIA~\cite{wang2024your}      & ECCV'24   & 20.36 & \underline{0.160} & \underline{0.758} & 286.3 & 19.91 & 0.205 & 0.64  & 313.3 \\
Follow-Your-Canvas~\cite{chen2024follow}          & AAAI'25   & 20.80 & \underline{0.160} & 0.726 & 242.8 & \underline{21.01} & \underline{0.178} & 0.699 & \underline{260.1} \\
Li et al.~\cite{lidynamic}                        & Neurips'25   & \underline{20.81} & 0.184 & 0.725 & \underline{234.7} & - & - & - & - \\
OutDreamer~\cite{zhong2025outdreamer}          & Arxiv'25   & 20.30 & 0.174 & 0.757 & 268.9 & - & - & - & -\\
GlobalPaint~\cite{pan2026globalpaint}          & Arxiv'26   & 20.91 & 0.154 & 0.762 & 227.8 & - & - & - & -\\

\midrule
\textbf{Ours}                      &           & \textbf{21.95} & \textbf{0.141}& \textbf{0.783} & \textbf{218.8} & \textbf{21.89}  & \textbf{0.143} & \textbf{0.783} & \textbf{242.8} \\
\bottomrule
\end{tabular}
\label{table1}
\vspace{-0.4cm}
\end{table*}

\subsection{Ablation Study}
\begin{figure}[htbp]
\centering
\includegraphics[width=1\linewidth]{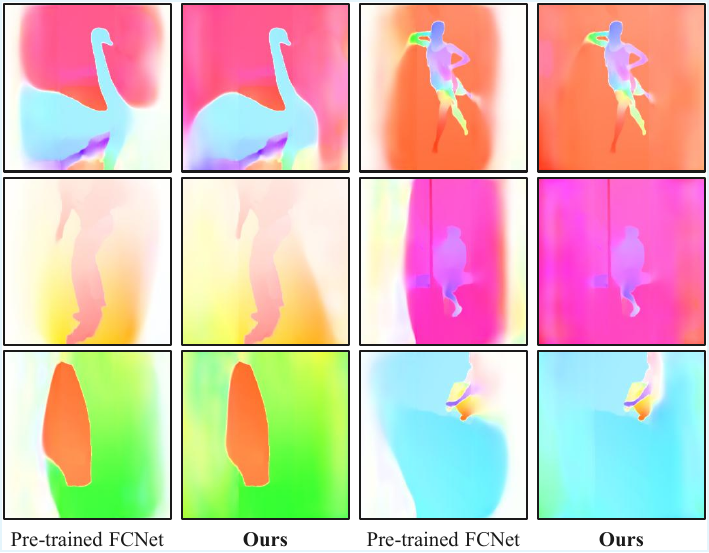}
\vspace{-7mm}
\caption{Comparison of flow completion results between an inpainting pre-trained model and our outpainting fine-tuned model.}
\vspace{-4mm}
\label{flow_completion}
\end{figure}

To validate the design of Seen-to-Scene, we conduct an ablation study on the DAVIS dataset, as summarized in Table~\ref{tab:ablation_components}. Each component progressively improves spatio-temporal coherence and overall visual quality. The baseline, which fine-tunes only the pre-trained video diffusion model, maintains temporal continuity but struggles to preserve the source content, resulting in inconsistent scene layouts in the outpainting regions. Introducing latent propagation provides an explicit mechanism for preserving source content by propagating structural cues from the observed regions. Even when using a pre-trained flow completion network trained for video inpainting, this propagation significantly improves spatial consistency and scene layout stability across frames. When the flow completion network is further fine-tuned for video outpainting, the performance improves substantially, highlighting the impact of the domain gap on flow completion quality as illustrated in Fig.~\ref{flow_completion}. Finally, the refinement module yields additional performance gains by suppressing minor artifacts using appearance cues.

\begin{table}[t]
\centering
\small
\caption{
Ablation on model components. 
(a). Video diffusion with fine-tuning only; 
(b). (a) + inpainting completion network with latent propagation; 
(c). (a) + flow completion network fine-tuning; 
(d). (c) + refinement module (our complete model).
}
\vspace{-3mm}
\begin{tabular*}{\columnwidth}{@{\extracolsep{\fill}}lcccc@{}}
\toprule
Method & PSNR$\uparrow$ & LPIPS$\downarrow$ & SSIM$\uparrow$ & FVD$\downarrow$ \\
\midrule
(a) & 19.42 & 0.254 & 0.556 & 586.3 \\
(b) & 20.92 & 0.195 & 0.687 & 355.7 \\
(c) & 21.61 & 0.144 & 0.742 & 220.7 \\
(d) & 21.95 & 0.141 & 0.783 & 218.8 \\
\bottomrule
\end{tabular*}
\vspace{-6mm}
\label{tab:ablation_components}
\end{table}

\begin{figure}[htbp]
\centering
\includegraphics[width=1\linewidth]{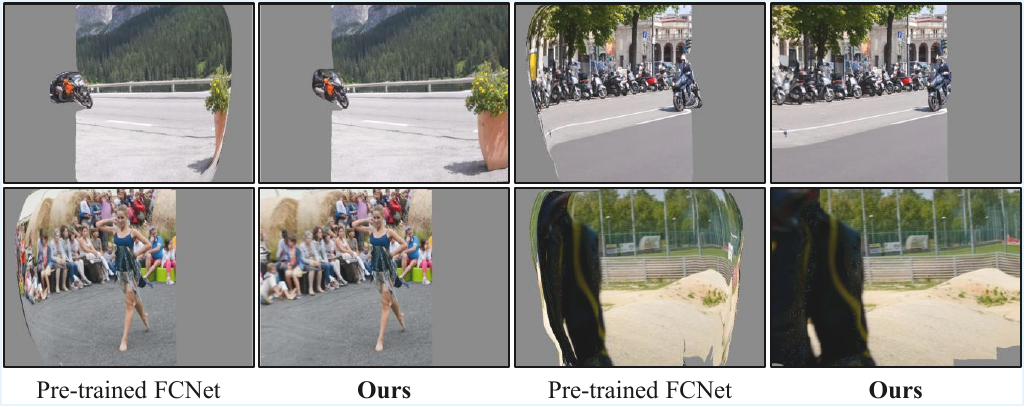}
\vspace{-6mm}
\caption{Visualization of warping results using an inpainting pre-trained and an outpainting fine-tuned flow completion network.}
\vspace{-5mm}
\label{propagated_result}
\end{figure}

\begin{figure*}[t]
\centering
\includegraphics[width=1\linewidth]{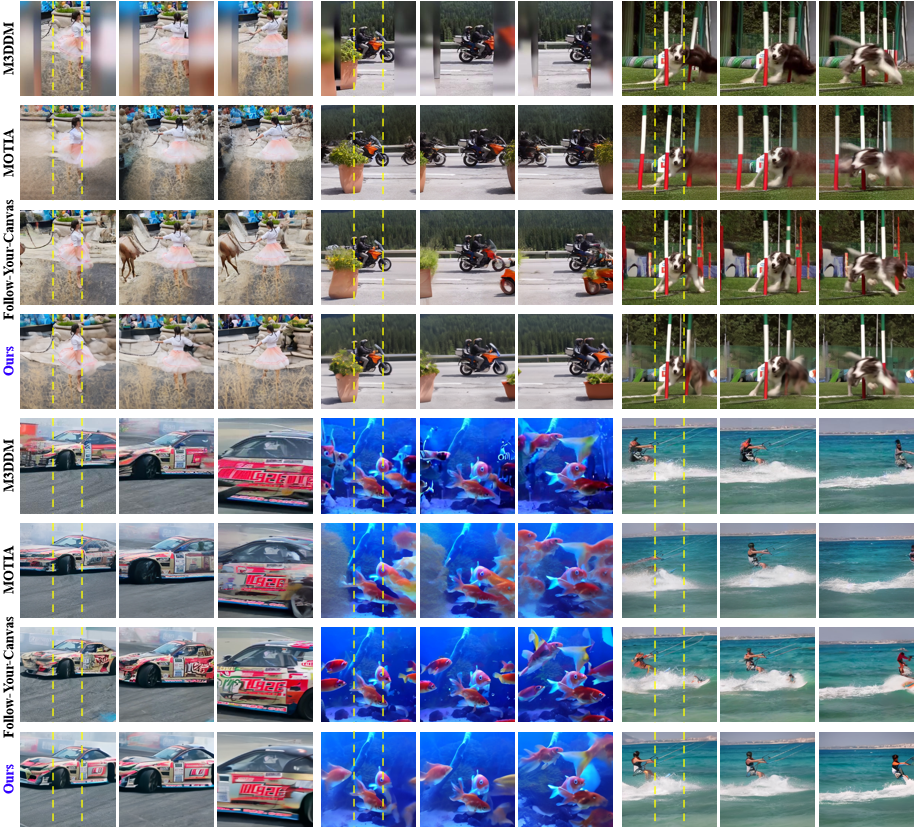}
\vspace{-6mm}
\caption{Qualitative comparison between Seen-to-Scene and State-of-the-art methods.}
\vspace{-4mm}
\label{figure4}
\end{figure*}

\noindent \textbf{Analysis of Domain Gap.}
To directly verify the necessity of fine-tuning the flow completion network for the video outpainting, we conduct a qualitative comparison using pixel-level warping results as illustrated in Fig.~\ref{propagated_result}. When using a pre-trained flow completion network trained for video inpainting, the completed flow often exhibits noticeable distortions in the outpainted regions, leading to unrealistic geometric deformations and inconsistent object boundaries. In contrast, after fine-tuning the flow completion network for video outpainting, our method produces more reliable flow fields, effectively preserving scene geometry and maintaining spatial consistency in the generated regions.

\vspace{-2mm}
\noindent \textbf{Analysis of Efficiency.} Seen-to-Scene also demonstrates strong computational efficiency. As illustrated in Fig.~\ref{fig:computational_cost} and Table~\ref{tab:inference_memory}, our latent-level propagation with a reference strategy significantly reduces both memory and computational overhead compared to pixel-level or sequential flow composition, resulting in faster inference.
\vspace{-3mm}

\begin{table}[htbp]
\centering
\small
\caption{Inference time and peak memory usage.}
\vspace{-3mm}
\begin{tabular}{lcc}
\toprule
Method & Inference time (s)$\downarrow$ & Peak memory (GB)$\downarrow$ \\
\midrule
M3DDM          & 43.127  & 9.63  \\
MOTIA          & 373.19  & 11.98 \\
\textbf{Ours}  & \textbf{12.109}  & \textbf{8.81}  \\
\bottomrule
\end{tabular}
\vspace{-3mm}
\label{tab:inference_memory}
\end{table}

\begin{figure}[t]
\centering
\includegraphics[width=0.95\linewidth]{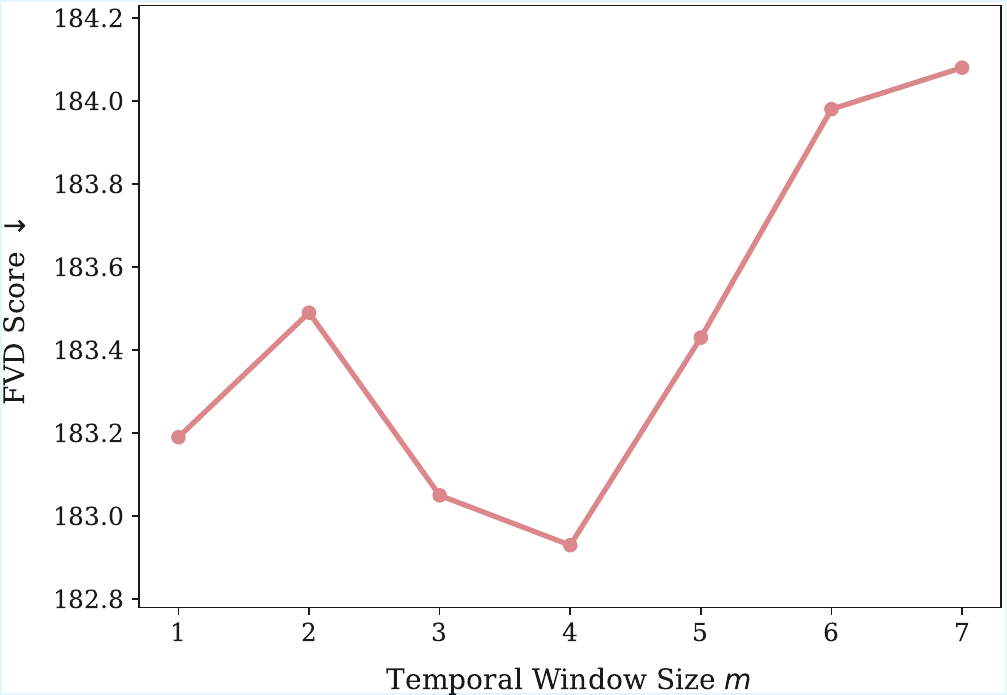}
\vspace{-3mm}
\caption{Effect of the temporal window size $m$ on FVD.}
\vspace{-6mm}
\label{temporal_window}
\end{figure}

\noindent \textbf{Analysis of Temporal Window Size.} We study the impact of the temporal window size $m$. As shown in Fig.~\ref{temporal_window}, we set $m=4$ in our method, which effectively balances temporal consistency and computational efficiency.

\section{Conclusion and Future Work}
\vspace{-5mm} In this work, we present Seen-to-Scene, a novel end-to-end framework that unifies propagation-based and diffusion-based paradigms for video outpainting. We further introduce an efficient and effective reference-guided latent propagation scheme. By combining propagation-based and generation-based paradigms, our approach achieves spatially faithful and temporally consistent results across diverse scenarios. Extensive qualitative and quantitative results demonstrate that Seen-to-Scene significantly pushes the limits of zero-shot video outpainting. However, our method still has several limitations, including potential detail loss introduced by latent-space propagation and the reliance on implicitly fine-tuning the flow completion network. We hope our framework encourages future research toward more robust and scalable video outpainting.


\begin{center}
	\vspace{5mm}
	\textbf{\Large{Appendix}}
	\vspace{5mm}
\end{center}
\appendix
\section{Video Inpainting vs. Video Outpainting}
\vspace{-3mm}
\begin{figure}[h] 
\centering 
\includegraphics[width=1\linewidth]{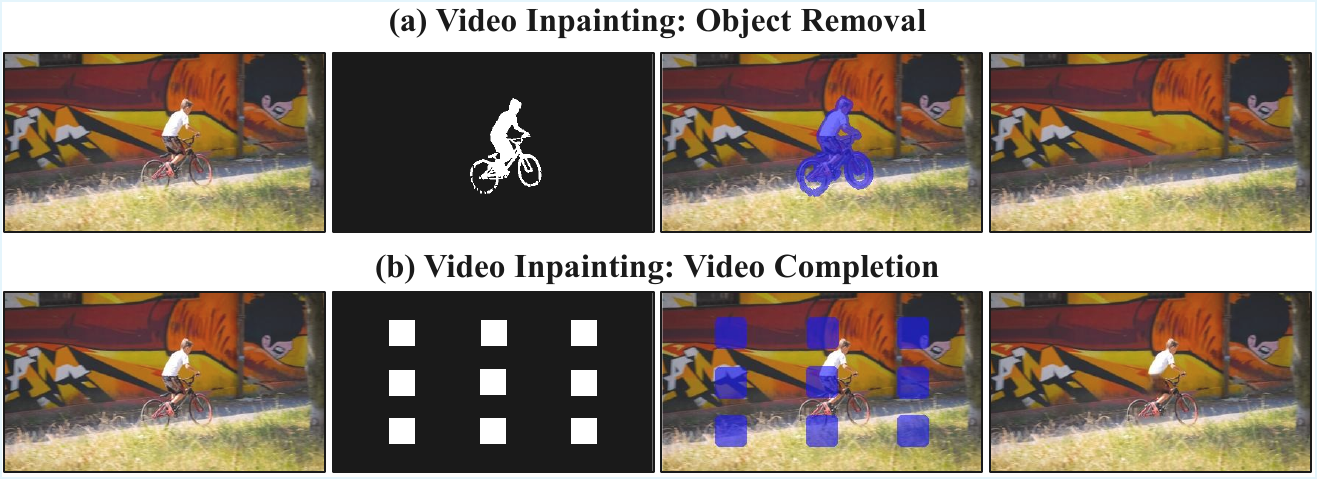} 
\vspace{-5mm} 
\caption{Two settings of video inpainting: (a) Object Removal eliminates dynamic objects and fills background regions; (b) Completion reconstructs locally missing areas within the visible frame.} 
\vspace{-6mm} 
\label{supple_figure_1} 
\end{figure}

\begin{figure}[h]
\centering
\includegraphics[width=1\linewidth]{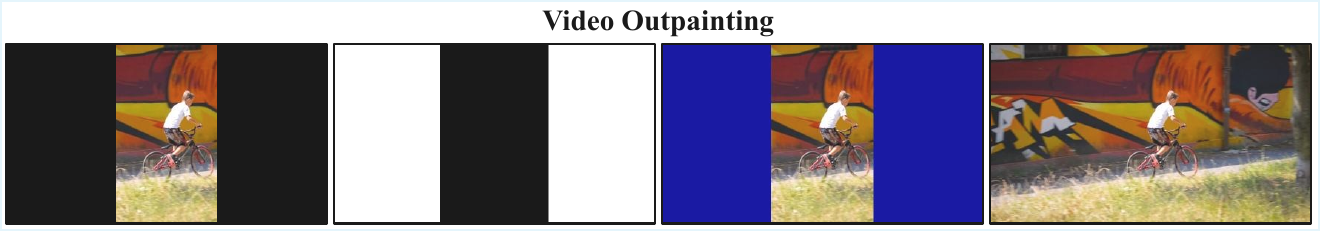}
\vspace{-5mm}
\caption{Conceptual illustration of video outpainting, which generates new regions beyond the original frame.}
\vspace{-3mm}
\label{supple_figure_2}
\end{figure}

\noindent \textbf{Conceptual Difference between VI and VO. }
\noindent As illustrated in Fig.~\ref{supple_figure_1}, video inpainting typically falls into two main settings: object removal and completion.
The object removal setting aims to eliminate dynamic foreground objects such as humans or vehicles and fill the resulting holes using motion-consistent background propagation. In contrast, the completion setting assumes corrupted or missing regions caused by occlusions, sensor noise, or editing, and seeks to reconstruct them based on surrounding spatio-temporal context. Both settings share a common characteristic: they operate strictly within the original frame boundaries, where abundant spatial and temporal information is available from the observed regions. As a result, the task primarily involves filling missing areas using nearby context rather than generating unseen content. Moreover, since the reconstruction is constrained to the given resolution, the computational complexity remains stable regardless of frame size, making video inpainting relatively contained in both scope and difficulty.

As illustrated in Fig.~\ref{supple_figure_2}, video outpainting fundamentally differs from inpainting in both objective and scope. Instead of restoring corrupted pixels within a given frame, it aims to extend the scene beyond the original field of view, synthesizing unseen structures and motion trajectories that remain unobserved in the input sequence.
This requires the model to infer plausible geometry, depth, and dynamics of the environment while maintaining consistency with the visible content. In other words, video outpainting transforms the problem from reconstruction to open-ended generation, demanding reasoning beyond directly observed data. Moreover, unlike inpainting, the computational cost of outpainting scales with the size of the expanded regions, as the model must process increasingly larger spatio-temporal domains during generation.

\begin{table}[t]
\centering
\small
\caption{Quantitative evaluation of ProPainter under the video outpainting setting.}
\vspace{-3mm}
\begin{tabular}{lcccc}
\toprule
Method & PSNR$\uparrow$ & LPIPS$\downarrow$ & SSIM$\uparrow$ & FVD$\downarrow$ \\
\midrule
ProPainter~\cite{zhou2023propainter} & 18.55 & 0.278 & 0.673 & 555.1 \\
\bottomrule
\end{tabular}
\vspace{-3mm}
\label{supple_table_propainter}
\end{table}

\begin{figure}[t]
\centering
\includegraphics[width=1\linewidth]{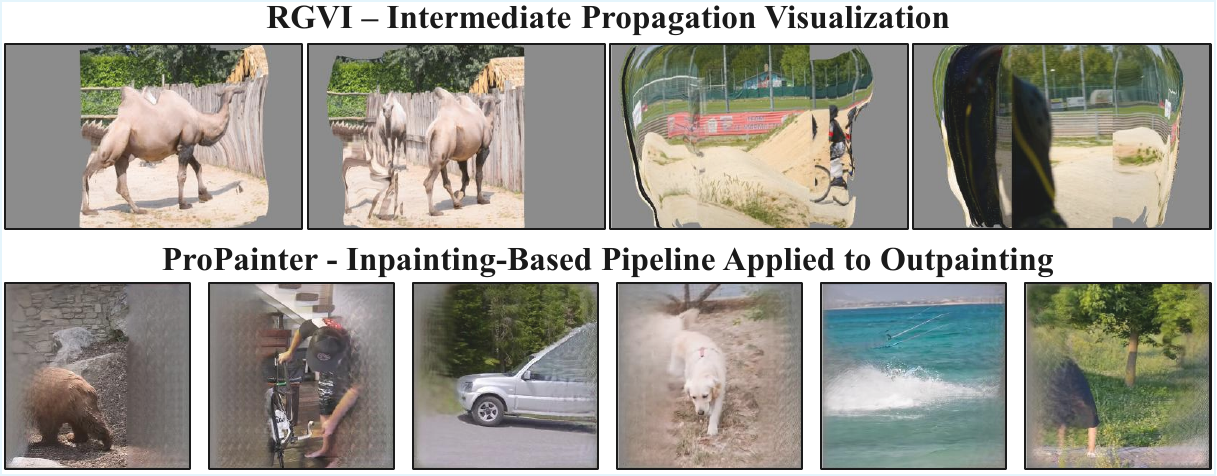}
\vspace{-5mm}
\caption{Limitations of Inpainting-Based Methods in Video Outpainting.}
\label{supple_figure_3}
\vspace{-3mm}
\end{figure}

\begin{figure*}[t]
\centering
\includegraphics[width=1\linewidth]{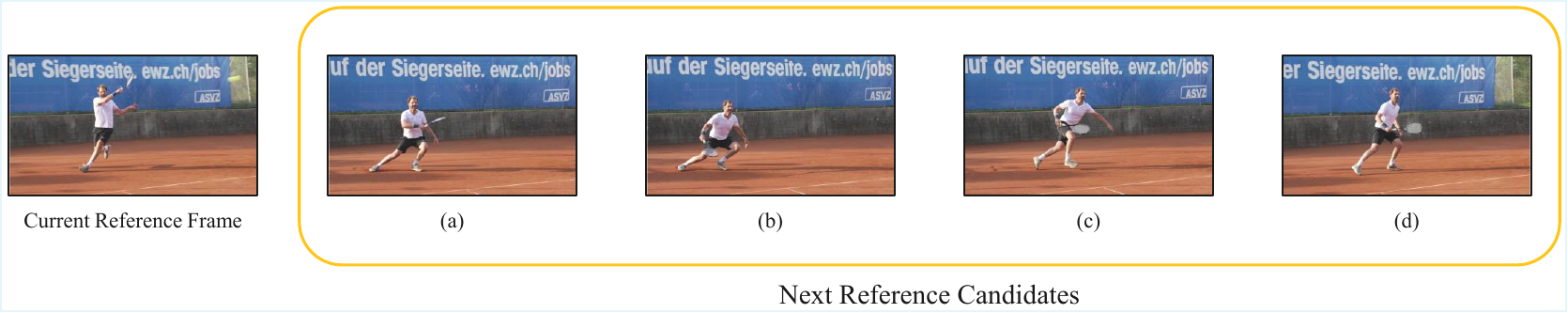}
\vspace{-7mm}
\caption{Comparison of fixed-stride and similarity-based reference selection. While the fixed-stride method selects frame (d) with redundant content, our method adaptively selects frame (a), which offers more informative cues for outpainting.}
\label{supple_figure_4}
\vspace{-5mm}
\end{figure*}

\vspace{1.5mm}
\noindent \textbf{Experimental Comparison Between VI and VO.} To validate the conceptual distinction between video inpainting and outpainting, we evaluate two representative inpainting-oriented propagation frameworks, RGVI~\cite{cho2025elevating} and ProPainter~\cite{zhou2023propainter}, under the outpainting setting. RGVI represents the current state of the art in flow-based propagation, while ProPainter is a widely adopted video inpainting model that integrates flow completion and content reconstruction into a unified pipeline. These models serve as strong baselines to analyze how inpainting-specific assumptions fail to generalize to the outpainting domain. As shown in Fig.~\ref{supple_figure_3}, we visualize the internal pixel propagation stage of RGVI, which propagates information across frames. In the outpainting scenario, motion must be extrapolated far beyond the observed boundaries, resulting in flow bleeding, distorted warping, and inaccurate motion prediction around scene edges and dynamic objects. These artifacts expose the limitations of inpainting-trained flow completion networks in handling unobserved regions. We further evaluate ProPainter by applying its full inpainting pipeline to the outpainting task. Although it performs well for inpainting, it fails to extend the scene over large unobserved regions. The model cannot generate new structures or motion beyond the original frame boundaries, resulting in spatially limited and incoherent generation. This indicates that inpainting-based reconstruction pipelines are inherently constrained by their reliance on visible context and struggle to generalize to large-scale scene expansion. Quantitative results under a 33\% horizontal outpainting setting further confirm these observations. Table~\ref{supple_table_propainter} reports significant degradation in both perceptual and temporal quality when ProPainter is applied to outpainting. The notably poor FVD score indicates severe temporal inconsistency, underscoring the fundamental gap between reconstruction-oriented inpainting and generation-oriented outpainting.

\vspace{1mm}
\noindent \textbf{Implications and Design Motivation.} 
These findings demonstrate that, although propagation mechanisms developed for video inpainting provide useful priors for maintaining temporal consistency, they cannot be directly applied to video outpainting due to fundamental differences in both domain and objective. In contrast, diffusion-based video outpainting methods primarily rely on generative synthesis. While fine-tuning large diffusion models can enhance perceptual realism, such approaches often overlook structural fidelity and fail to preserve alignment with the original scene layout. We argue that effective video outpainting requires integrating the complementary strengths of both paradigms—the structural reliability of propagation and the generative flexibility of diffusion—while mitigating their respective inefficiencies. To this end, we propose \textit{Seen-to-Scene}, an efficient propagation-guided diffusion framework that unifies structure propagation and content generation within a single diffusion process.

\vspace{2mm}
\section{Reference Frame Selection}

\noindent \textbf{Why reference frames are needed.} 
Video outpainting differs fundamentally from inpainting in that it must generate large outpainting regions while maintaining temporal and spatial coherence. To preserve structural consistency across frames, information from previously observed content should be propagated to the expanded regions. However, direct propagation from adjacent frames alone is often insufficient, since nearby frames tend to carry redundant information and contribute little to unseen areas. Moreover, propagating sequentially through all intermediate frames leads to excessive computational cost and accumulated warping errors. Therefore, selecting a small number of representative reference frames that can provide diverse structural and motion cues is crucial for efficient and reliable outpainting.

\begin{table*}[t]
\centering
\small
\caption{Quantitative evaluation and efficiency analysis of different window sizes for reference selection on the DAVIS dataset}
\vspace{-2mm}
\begin{tabular}{c|ccccc}
\toprule
Window Size ($w$) & PSNR$\uparrow$ & LPIPS$\downarrow$ & SSIM$\uparrow$ & FVD$\downarrow$ & Num. of Refs$\downarrow$ \\
\midrule
All Frames  & 21.65 & 0.151 & 0.767 & 183.19 & 70.65 \\
$w=2$       & 21.65 & 0.151 & 0.767 & 183.49 & 36.03 \\
$w=3$       & 21.65 & 0.151 & 0.767 & 183.05 & 26.46 \\
$w=4$       & 21.65 & 0.151 & 0.767 & \textbf{182.93} & 20.15 \\
$w=5$       & 21.65 & 0.151 & 0.767 & 183.43 & 15.92 \\
$w=6$       & 21.65 & 0.151 & 0.767 & 183.98 & 13.49 \\
$w=7$       & 21.65 & 0.151 & 0.767 & 184.08 & 11.65 \\
\bottomrule
\end{tabular}
\label{supple_table_window_selection}
\vspace{-6mm}
\end{table*}

\vspace{1mm}
\noindent \textbf{Reference Frame Selection vs Fixed Stride.} 
To evaluate the effectiveness of our reference selection strategy, we compare it against a fixed-stride selection scheme under identical video sequences. In the fixed-stride setting, reference frames are uniformly sampled at constant temporal intervals, regardless of scene dynamics. Such a strategy often leads to redundant reference choices, particularly when camera motion exhibits cyclic behavior (e.g., panning and returning to the same view). As a result, multiple references may contribute little new information for outpainting and may even fail to capture critical regions that should be propagated. In contrast, our similarity-based selection method adaptively identifies frames that provide complementary and non-redundant information relative to the target frame. This allows the propagation stage to effectively leverage structurally diverse cues from different temporal contexts, especially for expanding unseen regions. As illustrated in Fig.~\ref{supple_table_window_selection}, while the fixed-stride method selects frame (d) as the next reference, our method dynamically chooses frame (a), which contains more relevant and distinctive content for the outpainting area. This example demonstrates that our approach not only reduces redundant references but also enhances the informativeness and diversity of propagated content, leading to more consistent and complete video outpainting.

\vspace{2mm}
\noindent \textbf{Effect of window size on reference selection.}
We further investigate the impact of the temporal window size $w$ used during reference selection. This experiment aims to evaluate whether our adaptive reference selection can maintain comparable performance to full-frame propagation while offering significantly improved computational efficiency. We conduct the evaluation on the DAVIS dataset using 48-frame sequences and measure performance across various window sizes. For comparison, we also include a baseline where propagation is performed for all frames without reference selection. Table~\ref{supple_table_window_selection} shows the quantitative results under different window sizes. Across all settings, the PSNR, SSIM, and LPIPS values remain identical, indicating that spatial quality and perceptual similarity are not affected by the choice of window size. This consistency can be attributed to the design of our latent-space propagation. Since propagation occurs within the downsampled latent representation rather than at the pixel level, minor variations in reference selection have limited influence on pixel-domain metrics. Latents capture semantically aligned structure and motion cues that are robust to small temporal shifts, resulting in nearly identical spatial quality across different window configurations. However, the FVD scores exhibit slight fluctuations depending on $w$, reflecting the sensitivity of temporal dynamics to reference selection. The best FVD is obtained with $w=4$, which strikes an effective balance between temporal coverage and motion alignment accuracy. When $w$ is too small, reference diversity is reduced, increasing local flow accumulation errors. In contrast, excessively large windows include frames with greater motion variation, which introduces minor misalignment across long temporal gaps. This trade-off explains the gentle U-shaped trend in FVD as $w$ increases, confirming that our method achieves optimal temporal coherence at $w=4$.

\noindent \textbf{Efficiency of Reference selection.}
Across all DAVIS sequences, the average video length is approximately 70.65 frames. As shown in Table~\ref{supple_table_window_selection}, the number of selected references gradually decreases as the temporal window size $w$ increases. This indicates that our method performs efficient reference sampling without the need to propagate through every frame sequentially. Instead, it achieves the same temporal coverage using a significantly smaller set of key reference frames, reducing redundant propagation while maintaining the necessary contextual information for outpainting.

\section{Flow Completion Network (FCNet)}

\noindent \textbf{Domain gap between inpainting and outpainting.} 
Although flow completion networks trained for video inpainting can effectively estimate motion within observed regions, they exhibit notable instability when applied to video outpainting. This discrepancy stems from a fundamental domain gap between the two tasks. Inpainting assumes dense visual context within bounded regions, where the network can infer missing motion based on adjacent observations. In contrast, outpainting involves large unobserved areas beyond the original frame boundaries, requiring motion extrapolation into entirely unseen regions. Consequently, inpainting-trained flow completion models tend to produce unreliable predictions near boundary extensions, resulting in warped distortions, spatial discontinuities, and incomplete motion fields in the extrapolated zones. These artifacts propagate across frames and severely degrade temporal consistency in the generated videos.
 
\begin{figure*}[t]
\centering
\includegraphics[width=1\linewidth]{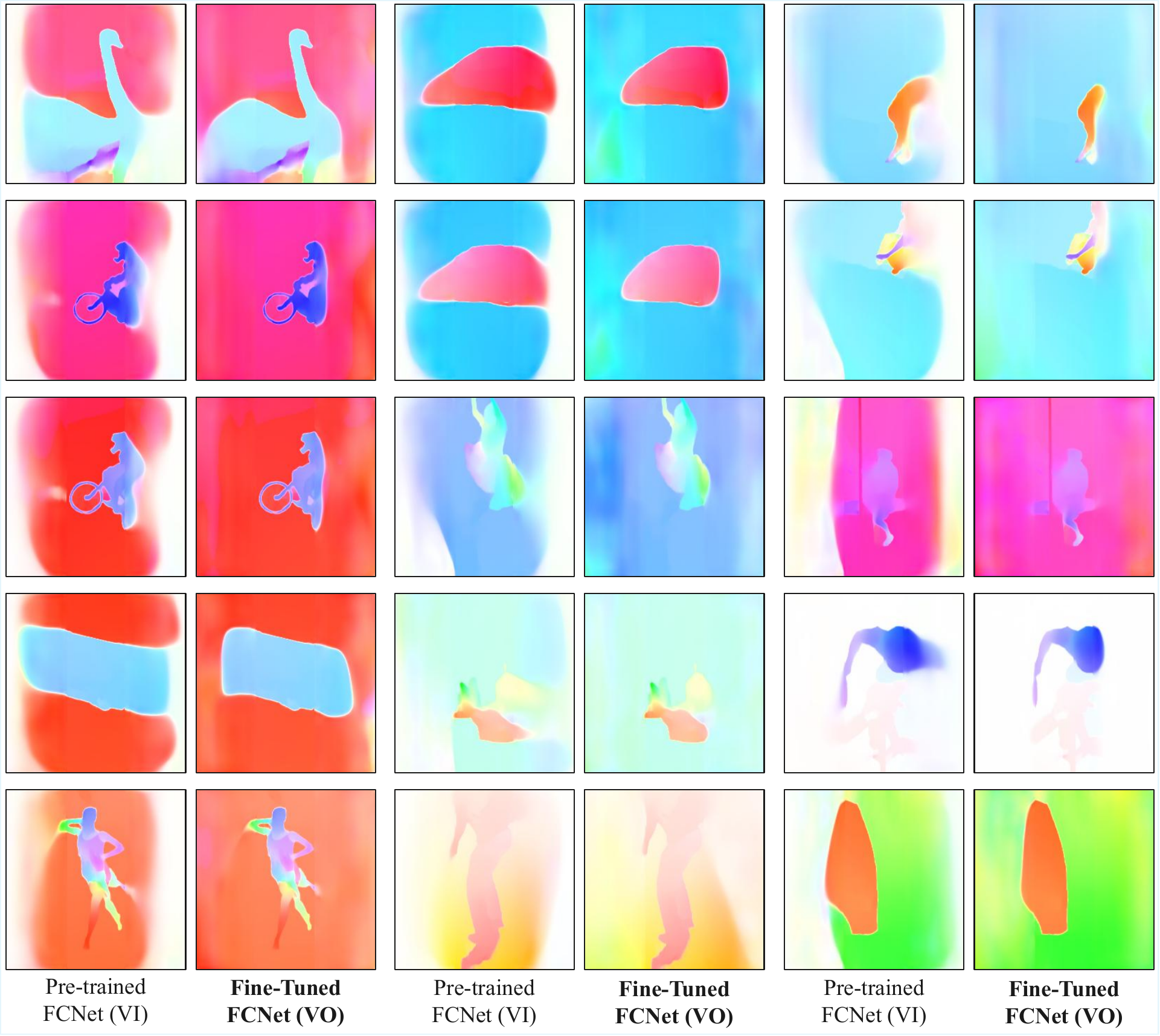}
\caption{\textbf{Visualization of flow completion results.} Our outpainting fine-tuned model produces geometrically stable and semantically separated motion fields compared to inpainting-pretrained flow completion.}
\label{supple_figure_6}
\end{figure*}

\begin{figure*}[t]
\centering
\includegraphics[width=1\linewidth]{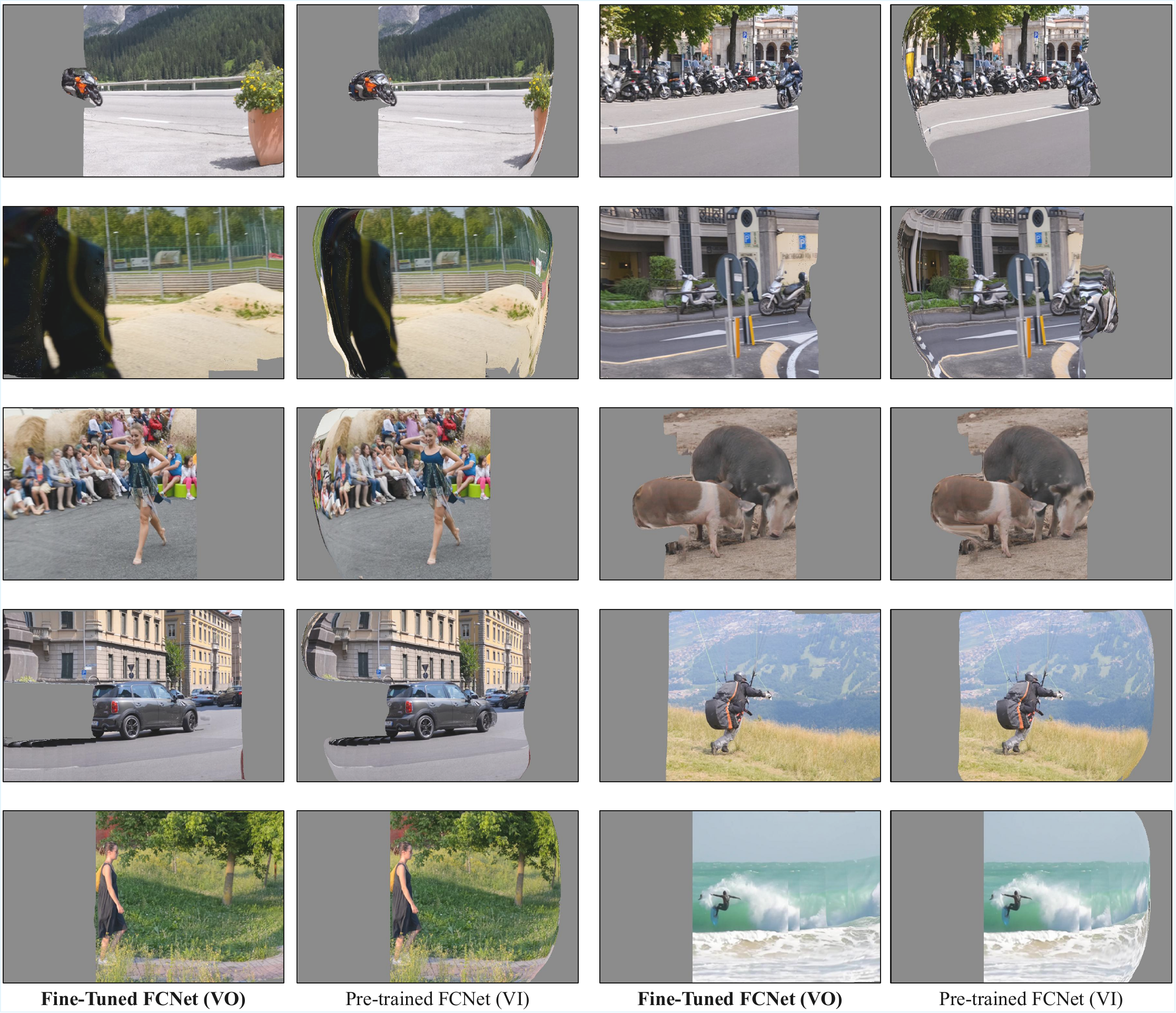}
\caption{\textbf{Visualization of pixel-level warping using different flow completion networks.} Our outpainting fine-tuned model achieves stable propagation, while the inpainting pre-trained model fails to propagate content effectively to outpainting regions.}
\vspace{-2mm}
\label{supple_figure_7}
\end{figure*}

\noindent \textbf{Flow Visualization and Analysis.}  
To further analyze the effect of domain adaptation, we visualize the flow completion results obtained from the inpainting pre-trained model and our outpainting fine-tuned version, as shown in Fig.~\ref{supple_figure_6}. For a fair comparison, both models are trained without reference frame selection and follow the same training configuration. It is important to note that the goal of our flow completion is not to hallucinate flow in unseen source regions, but to accurately complete motion in the areas necessary for warping during propagation. The objective is therefore to generate stable and geometrically consistent motion fields that enable reliable information transfer from source to outpainting regions. The inpainting pre-trained flow completion network exhibits severe \textit{flow bleeding} near boundary extensions or dynamic objects, where dynamic object motion incorrectly propagates into static background regions. In addition, its flow fields tend to rapidly lose magnitude toward the unobserved regions, indicating that the model struggles to infer motion continuity beyond visible boundaries. This limitation results in warped artifacts and incomplete propagation in the final outpainted frames. In contrast, our fine-tuned outpainting-specific flow completion network produces flow fields that are both spatially consistent and semantically aware of scene structure. Dynamic and static regions are clearly separated. Dynamic object motion is confined within its boundaries, and background flow remains stable without undesired bleeding. Moreover, the flow strength is maintained smoothly across extrapolated areas, allowing motion trajectories to extend naturally into the outpainting space. These results confirm that fine-tuning under the outpainting setting effectively adapts the flow completion process to handle large-scale scene expansion while preventing inter-region interference.

\vspace{1mm}
\noindent \textbf{Propagated Results Visualization and Analysis.} 
To directly verify the necessity of fine-tuning the flow completion network for the video outpainting domain, we conduct a qualitative comparison using pixel-level warping results, as illustrated in Fig.~\ref{supple_figure_7}.  Specifically, we propagate source content using two flow completion networks: (i) pre-trained for video inpainting and (ii) fine-tuned under the outpainting setting. For clarity of visualization, the warping is performed at the pixel level rather than in the latent space. When the inpainting-pretrained network is used, the propagation fails to extend accurately over large unobserved regions, resulting in incomplete motion transfer and significant spatial distortion near the extrapolated boundaries. This limitation arises because inpainting-trained models are optimized for local completion within visible regions and thus cannot extrapolate motion trajectories beyond the original field of view. As a consequence, even source-visible structures are often misaligned or missing in the propagated results, and the distortion becomes more pronounced as the distance from the original frame increases.  
Moreover, unstable flow estimation for dynamic objects leads to trajectory ghosting and flow bleeding artifacts, producing residual trails and inconsistent motion patterns. In contrast, when the flow completion network is fine-tuned on the video outpainting setting, the propagated content is spatially coherent and extends seamlessly into the expanded regions. The fine-tuned model learns to infer long-range and cross-boundary motion more robustly, resulting in stable propagation even for highly dynamic scenes. As shown in Fig.~\ref{supple_figure_7}, this adaptation effectively mitigates flow bleeding and geometric distortion, demonstrating that domain-specific fine-tuning is crucial for reliable motion estimation in video outpainting.

\subsection{Latent Propagation}

\begin{figure}[h]
\centering
\includegraphics[width=1\linewidth]{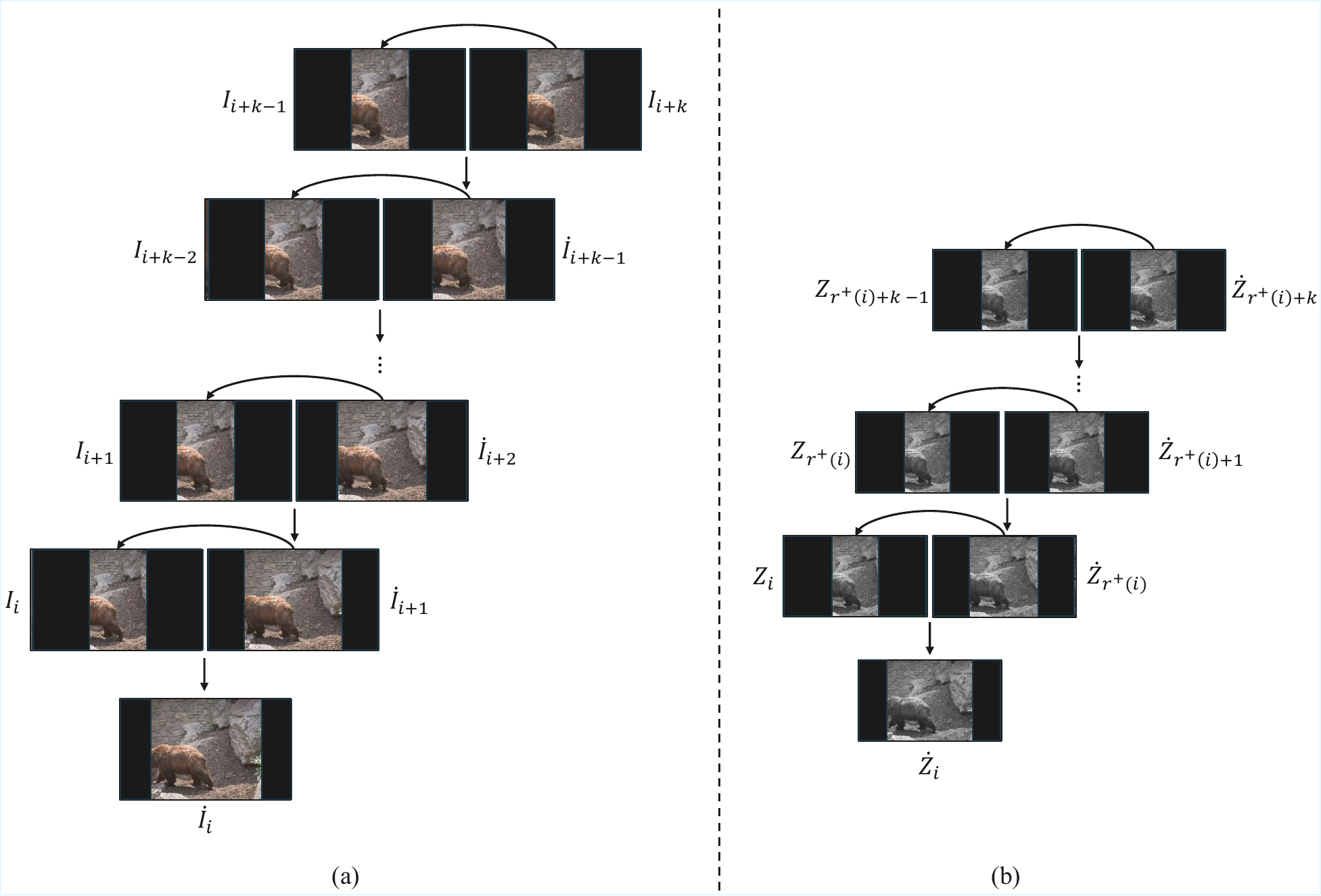}
\vspace{-5mm}
\caption{\textbf{Visualization of propagation strategies.} (a) Conventional sequential propagation accumulates flow across all intermediate frames, while (b) our reference-guided latent propagation performs efficient warping using only selected references in the latent space.}
\label{supple_figure_8}
\vspace{-3mm}
\end{figure}

\vspace{1mm}
\noindent \textbf{Reference-guided latent propagation.}
Instead of accumulating flow information across all intermediate frames, we utilize only a compact set of reference frames. Each reference frame is carefully selected to maximize informational diversity and to provide complementary content for the outpainting regions of the target frame. This design replaces dense sequential propagation with sparse, targeted correspondence, significantly reducing both temporal dependency length and the number of warping operations required.
Formally, the computational complexity is reduced from $\mathcal{O}(N^2)$ pairwise propagation to $\mathcal{O}(N \cdot R)$, where $R$ is the small number of reference frames ($R \ll N$). By restricting propagation to this compact set, redundant frame-to-frame accumulation is eliminated, which not only mitigates flow drift and error compounding but also leads to substantial improvements in runtime efficiency.

\subsection{Latent Refinement}
To refine warped multi-frame latents for video outpainting, we introduce a dual-branch refinement module that operates directly on the warped latent, the reference latent, and their associated binary masks. In the first branch, we concatenate the reference latent, the warped context latent, an outpainting mask, and a valid-warp mask, and use a lightweight CNN to predict spatially varying offsets and modulation masks for a deformable convolution, enabling the network to adaptively resample informative neighbors while implicitly correcting local misalignments in latent space. In parallel, a second branch processes the reference latent, warped latent, and outpainting mask using standard convolutions to produce a complementary refinement signal that emphasizes appearance and structural consistency around the outpainted regions. The outputs of the two branches are summed and passed through an additional convolutional layer to yield a refined latent, allowing the network to learn how to locally fuse deformation-aware and content-aware evidence. While we employ predicted optical flow to obtain the initial warped latents, our refinement module deliberately avoids using flow at this stage; instead, it learns offsets directly in latent space conditioned on visibility and outpainting masks. This design choice contrasts with prior refinement strategies that explicitly rely on optical flow fields, and is particularly important in the outpainting setting, where the predicted flow must be extrapolated into unobserved regions, making flow-based refinement susceptible to residual misalignments and artifacts. For bi-directional propagation, we apply the same refinement module independently to the forward-propagated and backward-propagated latents obtained from bi-directional warping, and then concatenate the two refined latents and pass them through a final convolutional layer to produce a unified, refined bi-directional propagation latent. By decoupling refinement from explicit flow and aggregating forward and backward evidence through content-adaptive latent offsets, our approach mitigates the propagation of errors from imperfect flow and yields propagated latents that are structurally well aligned between the source content and the bi-directionally warped regions.

\begin{figure}[t]
\centering
\includegraphics[width=1\linewidth]{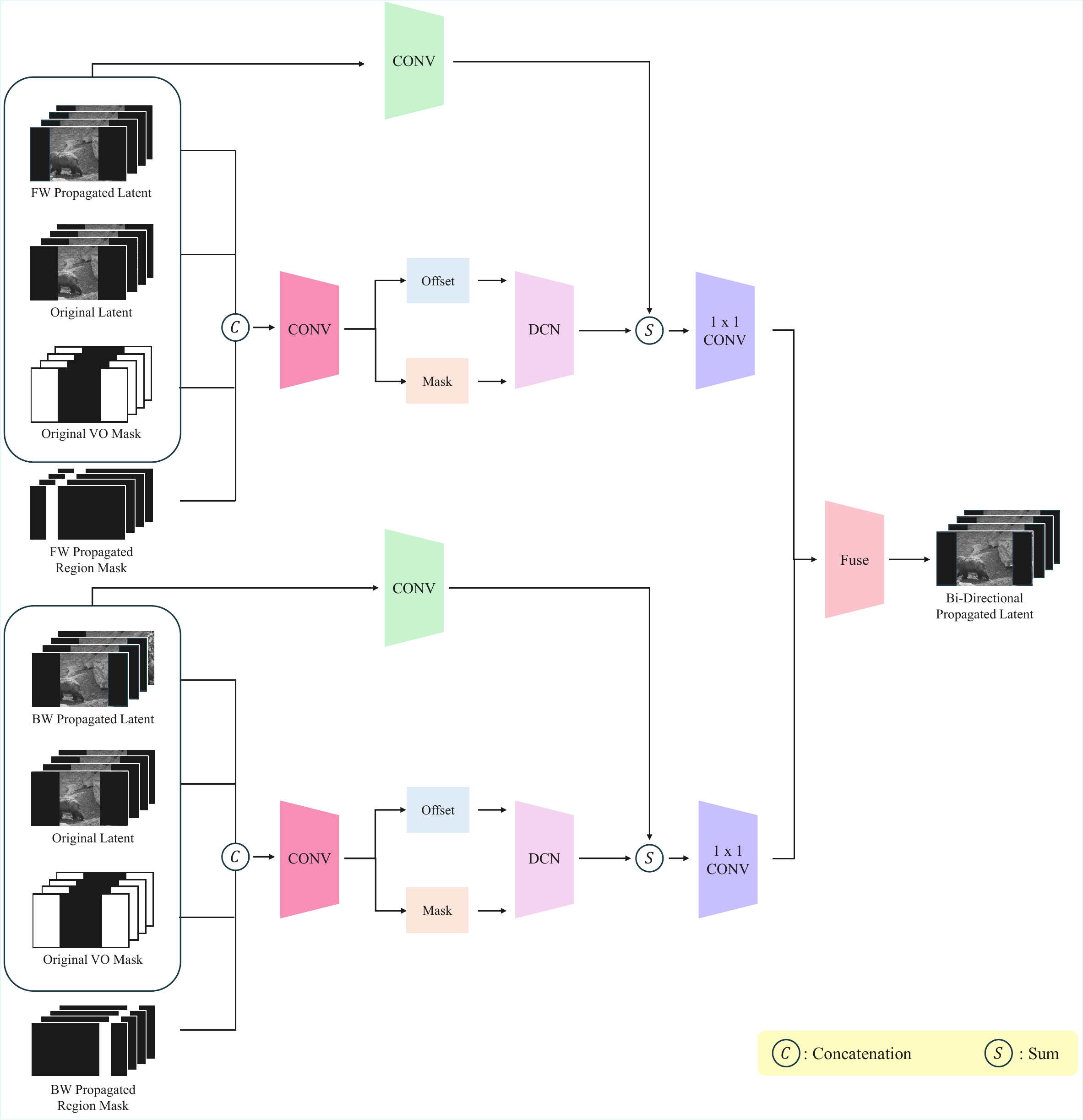}
\vspace{-5mm}
\caption{\textbf{Architecture of the latent refinement module.} It refines propagated latent features through spatial alignment and residual correction, enhancing temporal consistency and visual coherence in the outpainted video.}
\label{supple_figure_9}
\vspace{-2mm}
\end{figure}

\subsection{Pseudo Code for Latent Propagation}

\begin{algorithm}[t]
\caption{Reference-Guided Latent Propagation}
\label{alg:latent_prop}
\begin{algorithmic}[1]
\State \textbf{Input:} Video Sequence $\{I_0, I_1, \ldots, I_{T-1}\}$, \\ 
\ \ \ \ \ \ \ \ \ \ \ \ Mask Sequence $\{M_0, M_1, \ldots, M_{T-1}\}$, \\
\ \ \ \ \ \ \ \ \ \ \ \ Window size $w$
\State \textbf{Output:} Latent Sequence $\{Z_0, Z_1, \ldots, Z_{T-1}\}$
\vspace{3mm}

\State \textbf{Initialize:} $\mathcal{R} = \{I_0\}$, $c = 0$
\vspace{3mm}

\While{$c < T-1$}
    \State Define $\mathcal{W} = \{I_{c+1}, \ldots, I_{c+w}\}$
    \State Select $r = \arg\min_{I_j \in \mathcal{W}} \text{SSIM}(\text{g}(I_c), \text{g}(I_j))$
    \State Append $r$ to $\mathcal{R}$; \quad $c \gets r$
\EndWhile

\vspace{3mm}
\State \textbf{Flow Extraction and Completion}
\vspace{3mm}

\For{$i = 0, \ldots, T-1$}
    \For{each reference frame $r > i$}
        \State Accumulate flow $F_{r \rightarrow i} = F_{r \rightarrow r-1} \circ \cdots \circ F_{i+1 \rightarrow i}$
        \State Propagate latent feature $z_{r}$ via $F_{r \rightarrow i}$ to $z_{i}$
        \vspace{1mm}
        \State Propagate mask $m_{r}$ via $F_{r \rightarrow i}$ to $m_{i}$
        \vspace{1mm}
        \State $z_{i}^{aligned} = \mathcal{A}(z_{i}^{orig}, z_{i}^{prop}, m_{i}^{orig}, m_{i}^{prop})$
    \EndFor
    \vspace{1mm}
    \For{each reference frame $r < i$}
        \State Accumulate flow $F_{r \rightarrow i} = F_{r \rightarrow r-1} \circ \cdots \circ F_{i+1 \rightarrow i}$
        \State Propagate latent feature $z_{r}$ via $F_{r \rightarrow i}$ to $z_{i}$
        \vspace{1mm}
        \State Propagate mask $m_{r}$ via $F_{r \rightarrow i}$ to $m_{i}$
        \vspace{1mm}
        \State $z_{i}^{aligned} = \mathcal{A}(z_{i}^{orig}, z_{i}^{prop}, m_{i}^{orig}, m_{i}^{prop})$
    \EndFor
\EndFor
\vspace{1mm}
\For{$i = 0, \ldots, T-1$}
    \State $Z_i = \mathcal{F}(z_{i}^{aligned, FW}, z_{i}^{aligned, BW})$
\EndFor

\end{algorithmic}
\end{algorithm}

For clarity, we also provide the pseudo code in Algorithm~\ref{alg:latent_prop} summarizes the overall process of our Reference-Guided Latent Propagation. 
Here, $I$ denotes the input video frame sequence and $M$ represents the corresponding outpainting mask sequence that specifies outpainting regions to be generated. $\mathcal{W}$ denotes a local temporal window used to select reference candidates, and $\mathcal{R}$ indicates the selected reference frame set. $F$ represents the optical flow field between frames, where \textit{FW} and \textit{BW} correspond to forward and backward directions in the temporal domain, respectively. Each $z_i$ refers to a latent extracted from the corresponding frame, and $Z_i$ denotes the final bidirectionally fused latent feature of frame $i$. The alignment module $\mathcal{A}$ aligns propagated and original latent features using the propagated masks $m$, 
and $\mathcal{F}$ denotes the fusion operator that integrates forward and backward aligned features to produce temporally consistent latent representations.

\section{Details of Video Diffusion Models}

\subsection{Inference Stage}
During inference, we replace the ground-truth latent used in training with a randomly initialized Gaussian noise, drawn from a standard normal distribution, as the initial conditional latent. This noise serves as the starting point for generation, enabling the diffusion model to generate unseen content beyond the original frame boundary. To maintain structural and temporal coherence, the propagated latent features obtained from the reference-guided latent propagation are concatenated with the noise latent along the channel dimension. 
This concatenated representation provides both generative flexibility and spatial guidance, 
allowing the model to produce realistic and consistent outpainting content. No text prompts or additional conditioning signals are used during inference. Unlike prior diffusion-based approaches~\cite{fan2023hierarchical, wang2024your, chen2024follow}, we do not perform any fine-tuning, test-sample-specific optimization, complex inference strategy, or post-processing refinement. Our framework operates in a purely feed-forward manner, relying solely on the propagation-driven structural priors to achieve coherent and temporally stable video outpainting.

\subsection{Extension to Long Video Outpainting}
Sampling a long video sequence in a single diffusion pass requires excessive memory and restricts the effective temporal context available to the model. To alleviate this limitation, we utilize a sliding-window-based sampling strategy that enables scalable inference while preserving temporal consistency. Given a video of length $T$, we define a set of overlapping temporal windows $\mathcal{V} = \{ [s_i, e_i) \}_{i=1}^N$, where each window satisfies $e_i - s_i \le W$, and adjacent windows overlap with a stride $S$. At each diffusion timestep $t_k$, we apply the U-Net to every window independently and then average the per-frame noise predictions to obtain a unified global noise estimate:
\begin{equation}
\epsilon_t = \frac{1}{|\mathcal{V}(t)|} \sum_{i: t \in [s_i, e_i)} \epsilon_t^{(i)},
\end{equation}
where $\mathcal{V}(t)$ denotes the set of windows that include frame $t$. The averaged $\epsilon_{1:T}$ is then passed once to the scheduler’s update step to refine the entire latent sequence.

\section{YouTube-VOS Test Set}
To ensure transparency and reproducibility, we disclose the list of sequences used from the YouTube-VOS test set. Unlike prior works~\cite{fan2023hierarchical, wang2024your, chen2024follow, yu2025unboxed} that did not explicitly specify their evaluation subset, we randomly selected 60 sequences without any subjective bias to provide a fair and diverse benchmark for evaluating generalization performance. We release these sequence identifiers to facilitate future research, fair comparison, and reproducibility of our results.
\begin{verbatim}
0c7a4680db, 0d349f8286, 2e21c7e59b
2e129b0b09, 3b72dc1941, 3f2012d518
4b31a18d91, 4f5b3310e3, 5c3d2d3155
06a5dfb511, 6a75316e99, 6cced81d30
7daa6343e6, 8dea7458de, 9c4419eb12
13c3cea202, 24e2b52a4d, 37b4ec2e1a
37dc952545, 45fd60997a, 54ad024bb3
83a5056a16, 95ef69d827, 97b38cabcc
97fa40286c, 397dccb3a0, 459e70cd8e
03664dc880, 4035d3275c, 9787f452bf
40718bb478, 90949b2059, 547416bda1
607001c98f, 1320830fd2, 4348676053
6031809500, a9839ec6f2, ac4653b61d
b8bd20a472, b175cd8138, b492f67a89
b715879d5a, ba5dde67e9, c9ef04fe59
c16d9a4ade, cbea8f6bea, d1dd586cfd
d7ff44ea97, d59c093632, da9713ef3e
dab44991de, dc197289ef, e10236eb37
e1925724ab, eb49ce8027, f9eedb8691
f58212429d, fab725059c, fb104c286f
\end{verbatim}

\section{Additional Results}
\begin{figure*}[t]
\centering
\includegraphics[width=1\linewidth]{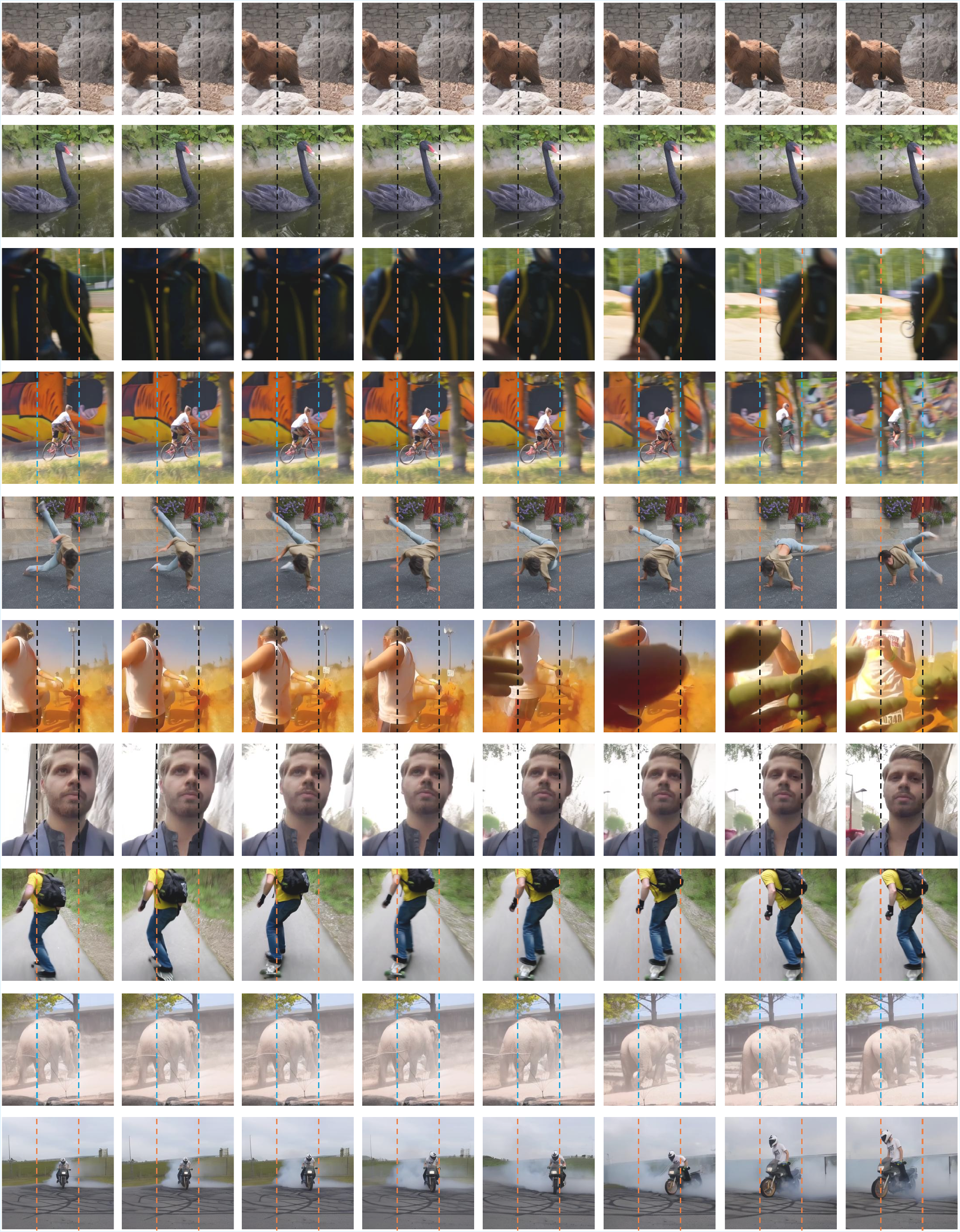}
\vspace{-8mm}
\caption{\textbf{Qualitative results of our method across diverse scenarios and environments.} The examples demonstrate the robustness and generalization capability of our model in handling various motion patterns, scene complexities, and outpainting configurations.}
\label{supple_figure_10}
\end{figure*}

To demonstrate the robustness and temporal coherence of our approach, we present multiple frame samples generated from various sequences under different scenarios in Fig.~\ref{supple_figure_10}. The results include both static and moving camera settings, single and multiple dynamic objects, and scenes containing humans, animals, and complex natural environments. Across these diverse cases, our method consistently produces temporally coherent and visually realistic outpainting results, preserving scene structure and motion continuity without noticeable flickering or spatial inconsistency. 

\begin{figure*}[t]
\centering
\includegraphics[width=1\linewidth]{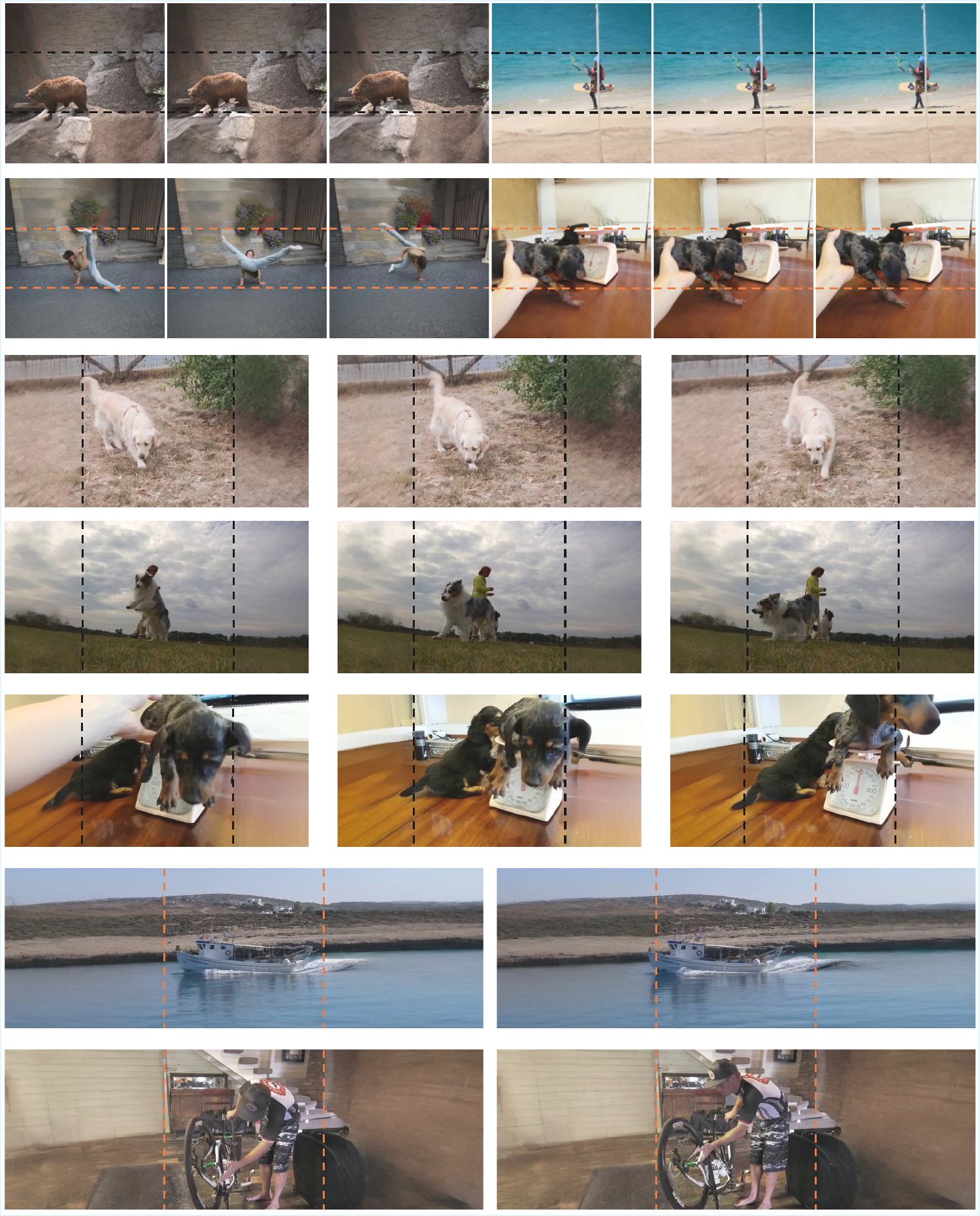}
\caption{\textbf{Qualitative results across different spatial resolutions and aspect ratios.} Our method maintains consistent visual quality and structure from $256\times256$ to extended formats such as $256\times512$ and $256\times768$.}
\label{supple_figure_11}
\end{figure*}

We further visualize our approach under varying output resolutions, aspect ratios and outpainting direction. As shown in Figure \ref{supple_figure_11}, our model effectively adapts to spatially expanded canvases, preserving coherent scene geometry and temporal consistency even at wide outpainting ratios. These results demonstrate that the proposed framework generalizes well beyond the training resolution, producing stable and high-quality content across diverse spatial configurations.

\vspace{1mm}
\section{Supplementary Video Demonstrations.}
We additionally provide demo videos in the supplementary materials, showcasing qualitative comparisons between our method and other existing approaches under identical experimental conditions.
Our results exhibit realistic visual synthesis with superior temporal consistency and structural coherence, while effectively suppressing hallucination artifacts across diverse scenarios.

\newpage
\vspace{2mm}
\noindent \textbf{Acknowledgement}. This work was supported by the National Research Foundation of Korea (NRF) grant funded by the Korea government (MSIT)(No. RS-2024-00340745) and Korea Institute of Science and Technology (KIST) Institutional Program (26E0062).
{
    \small
    \bibliographystyle{ieeenat_fullname}
    \bibliography{main}

@String(ECCV= {Eur. Conf. Comput. Vis.})

@String(AAAI = {AAAI})

@String(ECCV  = {ECCV})

@article{ho2020denoising,
  title={Denoising diffusion probabilistic models},
  author={Ho, Jonathan and Jain, Ajay and Abbeel, Pieter},
  journal={Advances in neural information processing systems},
  volume={33},
  pages={6840--6851},
  year={2020}
}

@inproceedings{dehan2022complete,
  title={Complete and temporally consistent video outpainting},
  author={Dehan, Lo{\"\i}c and Van Ranst, Wiebe and Vandewalle, Patrick and Goedem{\'e}, Toon},
  booktitle={Proceedings of the IEEE/CVF Conference on Computer Vision and Pattern Recognition},
  pages={687--695},
  year={2022}
}

@inproceedings{li2022towards,
  title={Towards an end-to-end framework for flow-guided video inpainting},
  author={Li, Zhen and Lu, Cheng-Ze and Qin, Jianhua and Guo, Chun-Le and Cheng, Ming-Ming},
  booktitle={Proceedings of the IEEE/CVF conference on computer vision and pattern recognition},
  pages={17562--17571},
  year={2022}
}

@inproceedings{zhou2023propainter,
  title={Propainter: Improving propagation and transformer for video inpainting},
  author={Zhou, Shangchen and Li, Chongyi and Chan, Kelvin CK and Loy, Chen Change},
  booktitle={Proceedings of the IEEE/CVF international conference on computer vision},
  pages={10477--10486},
  year={2023}
}

@inproceedings{cho2025elevating,
  title={Elevating Flow-Guided Video Inpainting with Reference Generation},
  author={Cho, Suhwan and Oh, Seoung Wug and Lee, Sangyoun and Lee, Joon-Young},
  booktitle={Proceedings of the AAAI Conference on Artificial Intelligence},
  volume={39},
  number={3},
  pages={2527--2535},
  year={2025}
}

@inproceedings{fan2023hierarchical,
  title={Hierarchical masked 3d diffusion model for video outpainting},
  author={Fan, Fanda and Guo, Chaoxu and Gong, Litong and Wang, Biao and Ge, Tiezheng and Jiang, Yuning and Luo, Chunjie and Zhan, Jianfeng},
  booktitle={Proceedings of the 31st ACM International Conference on Multimedia},
  pages={7890--7900},
  year={2023}
}

@inproceedings{wang2024your,
  title={Be-your-outpainter: Mastering video outpainting through input-specific adaptation},
  author={Wang, Fu-Yun and Wu, Xiaoshi and Huang, Zhaoyang and Shi, Xiaoyu and Shen, Dazhong and Song, Guanglu and Liu, Yu and Li, Hongsheng},
  booktitle={European Conference on Computer Vision},
  pages={153--168},
  year={2024},
  organization={Springer}
}

@article{chen2024follow,
  title={Follow-your-canvas: Higher-resolution video outpainting with extensive content generation},
  author={Chen, Qihua and Ma, Yue and Wang, Hongfa and Yuan, Junkun and Zhao, Wenzhe and Tian, Qi and Wang, Hongmei and Min, Shaobo and Chen, Qifeng and Liu, Wei},
  journal={arXiv preprint arXiv:2409.01055},
  year={2024}
}

@article{blattmann2023stable,
  title={Stable video diffusion: Scaling latent video diffusion models to large datasets},
  author={Blattmann, Andreas and Dockhorn, Tim and Kulal, Sumith and Mendelevitch, Daniel and Kilian, Maciej and Lorenz, Dominik and Levi, Yam and English, Zion and Voleti, Vikram and Letts, Adam and others},
  journal={arXiv preprint arXiv:2311.15127},
  year={2023}
}

@inproceedings{perazzi2016benchmark,
  title={A benchmark dataset and evaluation methodology for video object segmentation},
  author={Perazzi, Federico and Pont-Tuset, Jordi and McWilliams, Brian and Van Gool, Luc and Gross, Markus and Sorkine-Hornung, Alexander},
  booktitle={Proceedings of the IEEE conference on computer vision and pattern recognition},
  pages={724--732},
  year={2016}
}

@inproceedings{xu2018youtube,
  title={Youtube-vos: Sequence-to-sequence video object segmentation},
  author={Xu, Ning and Yang, Linjie and Fan, Yuchen and Yang, Jianchao and Yue, Dingcheng and Liang, Yuchen and Price, Brian and Cohen, Scott and Huang, Thomas},
  booktitle={Proceedings of the European conference on computer vision (ECCV)},
  pages={585--601},
  year={2018}
}

@inproceedings{xu2019deep,
  title={Deep flow-guided video inpainting},
  author={Xu, Rui and Li, Xiaoxiao and Zhou, Bolei and Loy, Chen Change},
  booktitle={Proceedings of the IEEE/CVF conference on computer vision and pattern recognition},
  pages={3723--3732},
  year={2019}
}

@inproceedings{yu2025unboxed,
  title={Unboxed: Geometrically and Temporally Consistent Video Outpainting},
  author={Yu, Zhongrui and Megaro-Boldini, Martina and Sumner, Robert W and Djelouah, Abdelaziz},
  booktitle={Proceedings of the Computer Vision and Pattern Recognition Conference},
  pages={7309--7319},
  year={2025}
}

@article{kingma2013auto,
  title={Auto-encoding variational bayes},
  author={Kingma, Diederik P and Welling, Max},
  journal={arXiv preprint arXiv:1312.6114},
  year={2013}
}

@inproceedings{rombach2022high,
  title={High-resolution image synthesis with latent diffusion models},
  author={Rombach, Robin and Blattmann, Andreas and Lorenz, Dominik and Esser, Patrick and Ommer, Bj{\"o}rn},
  booktitle={Proceedings of the IEEE/CVF conference on computer vision and pattern recognition},
  pages={10684--10695},
  year={2022}
}

@inproceedings{lee2025video,
  title={Video diffusion models are strong video inpainter},
  author={Lee, Minhyeok and Cho, Suhwan and Shin, Chajin and Lee, Jungho and Yang, Sunghun and Lee, Sangyoun},
  booktitle={Proceedings of the AAAI Conference on Artificial Intelligence},
  volume={39},
  number={4},
  pages={4526--4533},
  year={2025}
}

@article{wang2004image,
  title={Image quality assessment: from error visibility to structural similarity},
  author={Wang, Zhou and Bovik, Alan C and Sheikh, Hamid R and Simoncelli, Eero P},
  journal={IEEE transactions on image processing},
  volume={13},
  number={4},
  pages={600--612},
  year={2004},
  publisher={IEEE}
}

@inproceedings{teed2020raft,
  title={Raft: Recurrent all-pairs field transforms for optical flow},
  author={Teed, Zachary and Deng, Jia},
  booktitle={European conference on computer vision},
  pages={402--419},
  year={2020},
  organization={Springer}
}

@inproceedings{chen2024panda,
  title={Panda-70m: Captioning 70m videos with multiple cross-modality teachers},
  author={Chen, Tsai-Shien and Siarohin, Aliaksandr and Menapace, Willi and Deyneka, Ekaterina and Chao, Hsiang-wei and Jeon, Byung Eun and Fang, Yuwei and Lee, Hsin-Ying and Ren, Jian and Yang, Ming-Hsuan and others},
  booktitle={Proceedings of the IEEE/CVF Conference on Computer Vision and Pattern Recognition},
  pages={13320--13331},
  year={2024}
}

@inproceedings{zhang2018unreasonable,
  title={The unreasonable effectiveness of deep features as a perceptual metric},
  author={Zhang, Richard and Isola, Phillip and Efros, Alexei A and Shechtman, Eli and Wang, Oliver},
  booktitle={Proceedings of the IEEE conference on computer vision and pattern recognition},
  pages={586--595},
  year={2018}
}

@article{unterthiner2018towards,
  title={Towards accurate generative models of video: A new metric \& challenges},
  author={Unterthiner, Thomas and Van Steenkiste, Sjoerd and Kurach, Karol and Marinier, Raphael and Michalski, Marcin and Gelly, Sylvain},
  journal={arXiv preprint arXiv:1812.01717},
  year={2018}
}

@article{kingma2014adam,
  title={Adam: A method for stochastic optimization},
  author={Kingma, Diederik P},
  journal={arXiv preprint arXiv:1412.6980},
  year={2014}
}

@inproceedings{zhang2022flow,
  title={Flow-guided transformer for video inpainting},
  author={Zhang, Kaidong and Fu, Jingjing and Liu, Dong},
  booktitle={European conference on computer vision},
  pages={74--90},
  year={2022},
  organization={Springer}
}

@inproceedings{zhang2022inertia,
  title={Inertia-guided flow completion and style fusion for video inpainting},
  author={Zhang, Kaidong and Fu, Jingjing and Liu, Dong},
  booktitle={Proceedings of the IEEE/CVF conference on computer vision and pattern recognition},
  pages={5982--5991},
  year={2022}
}

@article{murakawa2026m3ddm+,
  title={M3DDM+: An improved video outpainting by a modified masking strategy},
  author={Murakawa, Takuya and Fukuzawa, Takumi and Ding, Ning and Tamaki, Toru},
  journal={arXiv preprint arXiv:2601.11048},
  year={2026}
}

@article{pan2026globalpaint,
  title={GlobalPaint: Spatiotemporal Coherent Video Outpainting with Global Feature Guidance},
  author={Pan, Yueming and Feng, Ruoyu and Bao, Jianmin and Luo, Chong and Zheng, Nanning},
  journal={arXiv preprint arXiv:2601.06413},
  year={2026}
}

@article{zhong2025outdreamer,
  title={Outdreamer: Video outpainting with a diffusion transformer},
  author={Zhong, Linhao and Li, Fan and Huang, Yi and Liu, Jianzhuang and Pei, Renjing and Song, Fenglong},
  journal={arXiv preprint arXiv:2506.22298},
  year={2025}
}

@inproceedings{lidynamic,
  title={Dynamic Shadow Unveils Invisible Semantics for Video Outpainting},
  author={Li, Ruilin and Yu, Hang and Qiu, Jiayan},
  booktitle={The Thirty-ninth Annual Conference on Neural Information Processing Systems}
}
}
\end{document}